\documentclass[runningheads]{llncs}

\usepackage{eccv}
   
\usepackage{eccvabbrv}
\usepackage{graphicx}
\usepackage{booktabs}

\usepackage[accsupp]{axessibility}  %
\usepackage{amsmath}
\usepackage{amssymb}
\usepackage{mathtools}

\usepackage{url}
\usepackage{algorithm}
\usepackage{algorithmic}
\usepackage{subcaption}
\usepackage{tabularx}
\usepackage[most]{tcolorbox}
\usepackage{lipsum,booktabs}
\usepackage{cuted}   %
\usepackage{capt-of} %
\usepackage{tabularx}
\usepackage{booktabs}
\usepackage{adjustbox}
\usepackage{ dsfont }
\usepackage{epigraph}

\usepackage{amsthm}
\usepackage{amssymb}
\usepackage{amsmath}
\usepackage{booktabs}  
\usepackage[utf8]{inputenc} 
\usepackage{enumitem}
\usepackage{multirow}        %
\usepackage{graphicx}        %
\usepackage{colortbl}        %
\usepackage{pifont}          %
\usepackage{array}           %
\usepackage{xcolor}
\usepackage{newfloat}
\usepackage{listings}
\usepackage{wrapfig}

\usepackage{hhline}
\usepackage{ gensymb }
\usepackage{hyperref}
\definecolor{truepink}{HTML}{D81B60}

\setlist[enumerate]{topsep=1pt, itemsep=5pt, parsep=0pt, partopsep=0pt}

\usepackage[capitalize]{cleveref}
\crefname{section}{Sec.}{Secs.}
\Crefname{section}{Section}{Sections}
\crefname{table}{Tab.}{Tabs.}
\Crefname{table}{Table}{Tables}

\begin{document}

\title{Forget, Anticipate and Adapt: Test Time Training for Long Videos}

\author{Rajat Modi\inst{1} \and
Sebastian Noel\inst{1}
\and
Xin Liang\inst{1}
\and
Yogesh S. Rawat\inst{1}}

\authorrunning{R. Modi, S. Noel, X. Liang, Y.S. Rawat}

\institute{Institute of Artificial Intelligence, University of Central Florida\\
\email{rajatmodi62@gmail.com, yogesh@crcv.ucf.edu}\\
\url{https://rajatmodi62.github.io/2025/07/10/ffn/}}

\maketitle
\begin{abstract}
Test Time Training (TTT) is a mechanism in which a model adapts to an incoming test-sample by performing some self-supervised (SSL) task and updating its weights even during inference. This procedure does not require labels at test-time. This paper focuses on TTT for long-videos. A major concern with existing approaches is: 1) they perform TTT updates using a sliding window containing frames in the past, whose compute increases linearly with the size of window. This becomes computationally intractable when the videos are hours long. 2) TTT is performed even when temporally close frames look similar, thereby consuming a lot of compute. \\

We present the Frame Forgetting Network (FFN) that: 1) operates on only three frames within the sliding window, namely the frame that exits, the current frame and the frame after that. The model still manages to retain temporal context and work for hours long-videos; 2) mathematically define a `surprise' metric: how much `new information' the incoming frame contains with respect to the past seen frame. This facilitates determining how to modify the effective window size during TTT and constitutes the core mechanism of an adaptive windowing algorithm. Additionally, we curate a dataset EpicTours containing up to 3 hour long videos of walking city-tours, whereas earlier datasets on this problem were only 5 min long. We demonstrate FFN's empirical effectiveness on dense-segmentation, video classification tasks, generalization to depth-estimation, and multi-hour long videos. The project page can be found at \href{https://github.com/rajatmodi62/ffn}{https://github.com/rajatmodi62/ffn}.
\end{abstract}

\section{Introduction}

Typically, machine learning models \textit{only} update their weights during training, but \textit{freeze} them during inference\cite{oquab2023dinov2,caron2021dino}. However, as humans, we possess the ability to \textit{continuously} learn and \textit{adapt} to our ever-changing environment. Test Time Training (TTT) shows the promise of bringing a similar adaptive capability to artificial intelligence, allowing a model to continuously learn and update its weights \textit{even while testing}, entirely without the need for ground-truth labels.\\

 \noindent However, applying TTT to \textit{video processing} presents a unique set of challenges. Existing online distillation methods are based on a student-teacher setup \cite{mullapudi2018online}, with the teacher usually deployed on a remote server and the student on a local device. However, this is a bottleneck in \textit{offline} real-world scenarios, such as \textit{disaster-prone regions}, where server connectivity may be limited.\\

 \noindent Consequently, the key challenge lies in determining how to perform TTT on videos locally, on-device, and computationally cheaply. This is especially critical for dense computer vision tasks, such as segmentation, where each pixel matters, and should ideally be achieved utilizing a \textit{single} model.\\

 \noindent Historically, there have been two approaches towards adaptation: either (i) train a \textit{dedicated} video model specifically for the task, or (ii) start from a pre-trained image model and adapt it accordingly. Unfortunately, large-scale pre-training datasets for video remain \textit{limited}, and efforts to repurpose image models are inherently constrained by the absence of explicit temporal information. Furthermore, existing TTT methods for videos rely on a \textit{sliding window approach}\cite{wang2025test}.  A sliding window serves as a \textit{fixed-size temporal buffer} that holds a defined number of preceding frames and moves forward \textit{step by step} as the video advances. Because the model re-evaluates all frames currently inside this window, it performs a \textit{duplicate}, highly expensive computation at every \textit{single} timestep. This \textit{redundant} computation is far \textit{too prohibitive} for long videos, thereby limiting practical deployment\\

 \noindent To overcome these limitations, we propose the Frame Forgetting Network (FFN). Our approach is built on the core insight that as the sliding window progresses, only one new frame enters, and one frame exits. Thus, a model \textit{only needs to pay} the computational cost of processing these crucial frames \textit{instead of} recalculating the entire window for every time-step. \\

\noindent Our FFN consists of two core components: a Memory Restoration Mechanism (MRM) and an Adaptive Window Algorithm (AWA). The MRM allows the model to actively `forget' its adaptation on past frames \textit{exiting} the window and `restore' the model's original predicted features using a temporal module. Next, the AWA algorithm can dynamically `anticipate' when adaptation is required, improving the management of computational load. Through extensive experiments across $11$ datasets, we validate FFN's \textit{competitive performance} on dense segmentation tasks, \textit{strong generalization} on depth estimation, and ability to learn \textit{stable representations} during recurrent video-processing.

\section{Frame Forgetting Network}

First, we shall talk about our problem statement, cover preliminaries on TTT, and then discuss our Frame Forgetting Network (FFN).

\subsection{Problem statement}
\noindent Given a \textit{streaming} long-video $\mathcal{V}=\{x_1,x_2,x_t,...,x_n\}$, a model can access only the past frames $x_1,x_2,...,x_{t-1}$, the current frame $x_t$, but \textit{not} the future. The problem is focused on adaptation: for $x_{t}$, how does the model \textit{decide} when to `adapt' or not. This adaptation should help improve downstream performance and be cheaper to compute. In this work, we focus on adaptation using TTT, an instantiation that allows a model to even update its weights during inference. 
\subsection{Preliminaries}

TTT involves two phases, a) Training Time Training and b) Test Time Training. \\

\begin{figure}[h] %
  \centering
    \includegraphics[width=\textwidth]{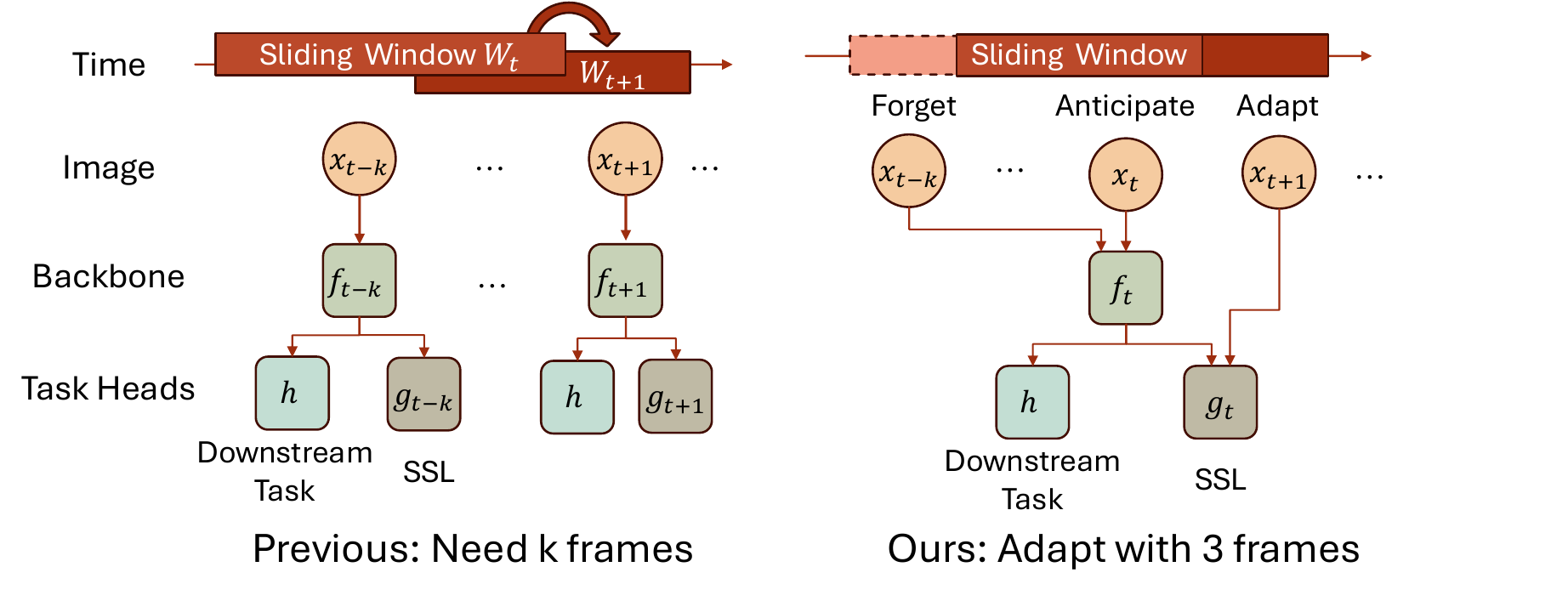} %
  \caption{(i) Test Time Training setup: Here $f$ is image-backbone, $g$ is SSL head, $h$ downstream head. Learning is self-supervised, where input test sample $x_{t}$ is transformed into $x_{t'}$ and compared again with $x_{t}$. (ii) TTT on videos relies on sliding windows, we denote two such windows $W_t$, $W_{t+1}$ . Notice that they contain a lot of overlapping frames, requiring N operations everytime. (iii) (Ours) We can perform TTT by gradually operating on only $3$ frames via a `forget, anticipate, adapt' procedure.}
  \vspace{-1em}
  \label{fig:teaser}
\end{figure}

\noindent \textbf{Training Time Training}: Fig\ref{fig:teaser}(i) illustrates how a TTT setup works. We consider a model containing backbone $f$, a downstream head $h$ and a self-supervised (SSL) head $g$ specializing in some task that does not require external labels (e.g., image reconstruction). First, we train $(f,g,h)$ \textit{jointly} in a train set, which consists of densely-annotated images. This first phase is known as \textit{training-time-training}\cite{wang2025test}.

\noindent \textbf{Test Time Training}: Here, the model is given access to a test-video consisting of several frames. The model looks at an incoming frame $x_t$ and tries to reconstruct it. As in Fig\ref{fig:teaser}, this involves feed-forwarding $x_t$ through the backbone $f$, then through the SSL-head $g$, thus computing the RGB reconstruction $x'_t$. Next, we estimate the reconstruction loss $\ell_{s}(x'_t, x_t)$, where $x'_t = g{\circ}f(x_t)$. This may be used to update $f,g$ over several iterations. The downstream head $h$ is \textit{kept frozen} throughout. This is known as \textit{test-time-training}.

\subsection{The Principle of locality}
\begin{figure}[h] %
  \centering
  \vspace{-1em}
    \includegraphics[width=\textwidth]{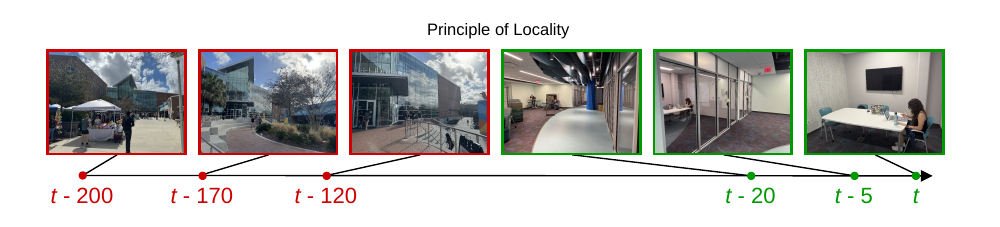} %
    \vspace{-2em}
  \caption{\textbf{The Principle of Locality}: The first $3$ frames (outdoors, marked in red) may not be directly relevant to last $3$ frames (indoors, marked in green), therefore during TTT, our model neglects them. Best viewed in color.}
  \vspace{-2em}
  \label{fig:locality}
\end{figure}

Adaptation in videos follows a `principle of locality'\cite{wang2025test}. Intuitively, a frame at $t=1$ (say indoors) \textit{may not} be relevant to the frame at $t=5k$ (say outdoors). Consider a current frame $x_{t}$. Instead of looking at the \textit{entire past timesteps}, existing TTT methods use a sliding window of size $k$, which we denote by $W_{t}$. More formally, the TTT update rule at $x_t$ is governed by:
\begin{equation}
\footnotesize
f_{t}, g_{t} = \arg \min_{f,g} \frac{1}{k} \sum_{t'=t-k+1}^{t} \ell_{s}(g \circ f(x_{t}), x_{t}),
\label{eq:ssl_update}
\vspace{1em}
\end{equation}
\noindent where $k$ is the size of a window containing frames encountered in the past (illustrated by window $W_t$ in Fig\ref{fig:teaser}). This creates two issues:

\begin{enumerate}

\item The window $W_t$ sums up the $k$ past frames for \textit{each} timestep. Inductively, $W_t$ slides over time (e.g., $W_{t+1}$) with the model's weights being reset after every update. The \textit{arg min} operator requires backpropagation through $f,g$ every timestep $t$. Thus, TTT quickly becomes computationally intractable, with a $2hr$ video ($7200$ frames) taking up to $8hrs$. 
\item The model wastes a lot of compute; for example, it shall perform TTT even when two successive frames are pixel-wise \textit{identical}.
\end{enumerate}

\noindent Note that $f,g$ can change their weights over time as the model undergoes adaptation. We denote this by the subscript $t$, namely $f_t$, and $g_t$.

\subsection{Method}

\noindent \textbf{The sliding window-invariant}: Consider a sliding window $W_{t} = [x_{t-k}, x_{t-k+1},
\\...,x_{t-1}, x_{t}]$, and the next window $W_{t+1}=[x_{t-k+1},...,x_{t},x_{t+1}]$. Mathematically, this reveals that \textit{both} have \textit{the same} number of frames \textit{except} for two: frame numbered $x_{t-k}$ that \textit{exits} the window $W_{t+1}$ and frame numbered $x_{t+1}$ that \textit{enters} the window $W_{t+1}$. \\

\noindent The intuition that operating on sliding windows requires a mechanism to update a running state and \textit{only} handle elements which come in/go out is well-known in data structures like arrays (e.g., Kadane's Algorithm \cite{bentley1984programming}). However, inducing such behavior is \textit{non-trivial} in neural nets, which in-principle may be treated as forms of parallel distributed memories\cite{hinton1986learning}. \\

\noindent This reveals that while making a transition from $W_{k}\rightarrow W_{k+1}$, we \textit{don't} need to process $k$ frames. Rather, we can operate on \textit{only} $3$ frames by defining three mechanisms:

\begin{enumerate}

\item \textit{Forget:} for frame $x_{t-k}$, `forget' the model's features \textit{after} adaptation, and `restore' the model's predicted features to those \textit{prior} to TTT. 

\item \textit{Anticipate:} The current frame $x_{t}$ under processing can be used to `anticipate' what shall come next at $x_{t+1}$. For brevity, we call this prediction $x'_{t+1}$

\item \textit{Adapt:} Compare $x'_{t+1}$ with the actual frame $x_{t+1}$. If the difference is greater than some threshold $\tau$, the model chooses to adapt on $x_{t}$. Otherwise, it can perform simple inference on $x_{t}$ and move on. Note that inference is equivalent to the feed-forward $f\circ h(x_t)$ through the model.
\end{enumerate}
\vspace{1em}

\noindent An initial approach might be to hard-code $\tau$ to some value (say $0.5$). However, this makes the model's behavior \textit{static}: there \textit{might be} some consecutive frames who have large pixel-difference, but do not deserve adaptation (for e.g., slight rotation of images of the same object). Thus, we need a mechanism that could govern this $\tau$ dynamically.\\

\noindent Inspired by this, we present the \underline{F}rame \underline{F}orgetting \underline{N}etwork (FFN). It consists of two components: 1) \underline{M}emory \underline{R}estoration \underline{M}echanism (MRM) 2) \underline{A}daptive \underline{W}indow \underline{A}lgorithm (AWA). MRM allows the model to `forget' the frame that moves out of the window ($x_{t-k}$), while AWA allows the model to \textit{anticipate} if $x_{t}$ deserves adaptation, by dynamically estimating $\tau$. Next, we explain these blocks.

\subsection{ The Memory Restoration Mechanism (MRM)}

\begin{figure}[t] %
  \centering
  \includegraphics[width=0.9\textwidth]{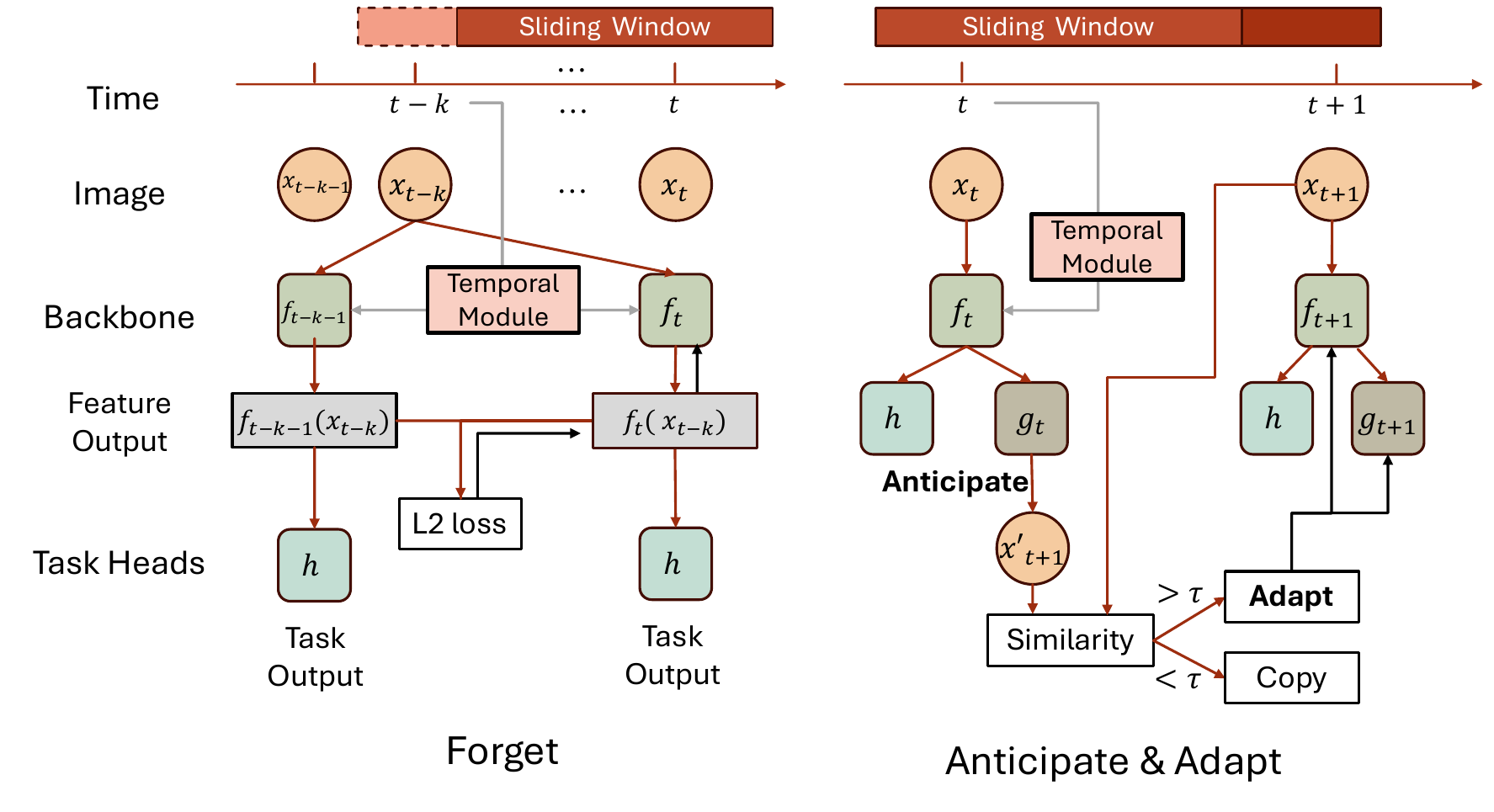} %
  \vspace{-1em}
  \caption{\textbf{Frame Forgetting Network}(i) Forget step: Backbone $f_{t}$ takes the frame $x_{t-k}$ to forget as input, along with timestep $t-k$, to get the feature encoding, $f_{t}(x_{t-k})$. To \textit{forget} its adaptation on $x_{t-k}$, backbone is trained with pre-adapted features $f_{t-k-1}(x_{t-k})$ via L2 loss. (ii) Anticipate and Adapt step: Given frame $x_{t}$ model predicts the next frame $x'_{t+1}$. This is compared with actual frame $x_{t+1}$ to make estimate whether to adapt or not. 
  }
  \vspace{-1em}
  \label{fig:teaser_wrap}
\end{figure}

\noindent To forget the model's adaptation on $x_{t-k}$, we can first \textit{cache} its features before adaptation; i.e., we estimate $f_{t-k-1}(x_{t-k})$. The idea is that the \textit{current} weights of the model $f_t$ should be adjusted so that it can predict the earlier features $f_{t-k-1}(x_{t-k})$. This suggests that the model $f$ requires two things: i) input $x_{t-k}$, and ii) a sense of time $t-k$. We propose instilling this via a \textbf{temporal module}. \\

\noindent We implement a temporal module as a three layered  MLP\footnote{This is because neural fields work well with coordinate information\cite{mildenhall2021nerf}}. First, the timestep $t-k$ corresponding to $x_{t-k}$ is injected into the temporal module. Instead of conditioning the temporal module by a single integer, we implement $t$ as a one-dimensional positional encoding, similar to the one used in transformers\cite{vaswani2017attention}. Note that injecting time $t$ in an image-based backbone $f$ may be cheaper than training a video-backbone from scratch. The output of temporal module is then used to condition the backbone $f_{t-k-1}$, resulting in two input arguments $x_{t-k},t-k$. Mathematically, we can minimize the following:
\vspace{-0.5em}
\begin{equation}
\ell = f_t(x_{t-k},t-k) - f_{t-k-1}(x_{t-k}, t-k)  
\end{equation}
\noindent This can be used to make a single gradient step through backpropagation and adjust the weights of the backbone $f$. 

\subsection{ Adaptive Window Algorithm (AWA)}

\noindent \textbf{An Anticipative Head:} Given the current frame $x_{t}$, we need to \textit{anticipate} whether adaptation is required. The idea is to allow the model to take $x_{t}$ as input, predict the next frame $x_{t+1}'$, and compare with the real frame $x_{t+1}$. The average of pixel-wise difference can then serve as an indicator if adaptation is required\footnote{This anticipative-head may be learned with as few as 1 long-video during the training phase of TTT. We also experimented with the \textit{k-step} future-frame predictor\cite{zhang2019lookahead,oord2018representation}, but found the \textit{next} timestep predictor to work quite well (similar to classical auto-regression in language models) }. Thus, we define the visual space difference as 
\begin{equation}
\footnotesize
    v_{visual}(t) = \frac{2}{\sqrt{h\times w}} \sum_{h,w} \frac{\|x_{t+1} - x_{t+1}'\|_2}{255}
\label{eq:v_visual}
\end{equation}
\noindent One concern about equation \ref{eq:v_visual} is that \textit{if} an incoming frame $x_{t+1}$ just `rotates' slightly over the current frame $x_{t}$, $v_{visual}$ shall be very high owing to the fluctuations in the \textit{pixel} space (for e.g., capturing a video over phone). Intuitively, however, these two frames still look almost the same. This means that \textit{low-level} information \textit{might} not be a reliable indicator. \\

\noindent Thus, we make use of the following observation: \textit{higher} layers of a model $f$ remain \textit{invariant} to minor-changes in input frames. Inspired by this, we define a memory buffer $B=[v_{latent}(t_1), v_{latent}(t_2), ...,v_{latent}(t_{last})]$, where $v_{latent}(t_i) = f_{t_{i}}(x_{t_i})$. Note that different $t_{i}$ here denote the timesteps when the model actually `adapted' in the past, and \textit{need not} be consecutive. The idea is that if the $latent$ feature of the \textit{most recently adapted} past frame ($t_{last})$ is very similar to the latent feature of the current frame $x_t$, we must \textit{not} adapt. Please note that $t_{last}$ need \textit{not} be equal to $t-1$. We define an anticipation matrix A as: 

\begin{equation}
    A = \frac{v_{latent}(t_i) \cdot v_{latent}({t_{last}})}{\|v_{latent}(t_i)\|_2 \|v_{latent}({t_{last}})\|_2}
\label{eq:f_latent}
\vspace{1em}
\end{equation}

\noindent \textbf{The surprise metric}: Both equations \ref{eq:v_visual} and \ref{eq:f_latent} can be unified together to define a metric that measures how much new information (surprise $S$) a frame $x_{t}$ provides to the model. We define $S$ as:
\vspace{1em}
\begin{equation}
\footnotesize
S_t = \left[ \log(1 + v_{\text{visual}}(t)) \right] \times [1 - A]
\vspace{1em}
\end{equation}
\noindent Next, we consider three cases:
(a) when video contains static frames (e.g., CCTV camera feed): Here $v_{visual}\approx 0, A\approx 1$, so surprise $S_t \approx 0$; (b) when video contains a shaking camera (e.g., person captures a video with a head-mounted display): Here the scene remains the same, but frames change a lot. Therefore $v_{visual}\uparrow$, $A\approx constant$, so surprise $S_t$ \textit{stays low }; (c) jump cuts, where the scene changes suddenly (say a transition from indoors to outdoors). There $v_{visual}\uparrow$, $A\uparrow$, however $v_{visual}$ increases \textit{far more} than $A$\footnote{ there are more numbers of pixels in $v_{visual}$ than the dimensions in $A$, so we multiply with the factor of $\frac{2}{\sqrt{N}}$ in Eq\ref{eq:v_visual} to compensate.}, so surprise $S_{t}$ takes on a \textit{very high} value. \\

\noindent Both cases b) and c) suggest that if $S_{t}$ exceeds or equals to \textit{some adaptive threshold} $\tau_t$, the model should perform TTT. Otherwise, it may perform inference on the frame $x_t$ (i.e., a single forward pass $h(f_{t-1}(x_t))$) and move on to the next frame $x_{t+1}$. In this case, $f_{t}=f_{t-1}$, since it is \textit{not} updated. Next,  we derive $\tau_t$.\\

\noindent \textbf{Deriving the adaptive threshold $\tau_t$:} Recall that the buffer $B$ contains latent features of all frames on which TTT was performed. We assume that the \textit{size} of this buffer is $N$. Each element in the buffer contains the latent feature $v_{latent}(t_i)$, and the surprise computed at $t_{i}$. We define $\tau_t$ as :
\begin{equation}
\mu_{t} = \frac{1}{W} \sum_{i=t-W}^{t-1} S_{i} \quad ; \quad \sigma_{t} = \sqrt{\frac{1}{W} \sum_{i=t-W}^{t-1} (S_{i} - \mu_{t})^{2}}
\label{eq:running_avg}
\end{equation}
\vspace{-1em}
\begin{equation}
\tau_{t} = \mu_{t} + \sigma_{t}
\end{equation}
Intuitively, $\mu_t$ tracks how the mean of surprise flows over time\cite{wei2018learning,layzer1975arrow}, $\sigma_t$ estimates variance, $\tau_t$ is one standard deviation away from the mean\footnote{Statistically, the central limit theorem says that the majority Gaussian probability mass lies in one standard deviation, which informed this choice.}. Note that while the size of buffer $N$ is fixed, the threshold $\tau_t$ itself is dynamic. Therefore, unlike equation \ref{eq:ssl_update} where TTT was done for \textit{each} iteration, the model now dynamically decides when to do TTT. Also, we are processing three frames in a timestep and \textit{not} N.\\

\noindent \textbf{Revisiting the cold-start problem:} Initially, our model does not see any frame, hence the buffer $B$ is empty. Therefore, it is difficult to make a reliable estimate of whether to adapt or not. This issue also appears as a cold-start problem in recommendation systems\cite{wei2017collaborative}. We attempt to resolve this by `adapting' $B$ initial-frames until the buffer is full. Since $W<<<$ video-length, this approach worked quite well, albeit at the expense of some initial-lag in the system\footnote{Buffer $B$ should contain 60 frames initially. Each frame takes $0.7sec$ for TTT, which incurs about  42 seconds of lag as $B$ gets initialized.}.

\section{Experiments on FFN}

Here, we experiment with FFN across a wide range of benchmarks, for example, (i) dense tasks like semantic, instance, and panoptic segmentation; (ii) generalization across video depth-estimation; (iii) action-classification; (iv) segmentation on multi-hour long videos. \\

\noindent \textbf{Datasets:} We report video segmentation results in COCO-Videos\cite{wang2025test}, KITTI-STEP\cite{weber2021step}. For action classification, we report UCF101\cite{ucf101}, Something-Something v2\cite{goyal2017something}. Similarly, for depth-estimation, we create a similar setup as Video-DepthAnything (CVPR'2025)\cite{yang2024depth}, and show results on $6$ datasets (KITTI\cite{geiger2013vision}, Scannet\cite{dai2017scannet}, Bonn\cite{palazzolo2019iros}, NYUv2\cite{Silberman:ECCV12}, Sintel\cite{Butler_Wulff_Stanley_Black_2012}). As shown in Table \ref{tab:dataset_details}, most of these datasets contain only videos up to $5 mins$ long.\\

\begin{figure}[t] %
  \centering
  \vspace{-1em}
    \includegraphics[width=\textwidth]{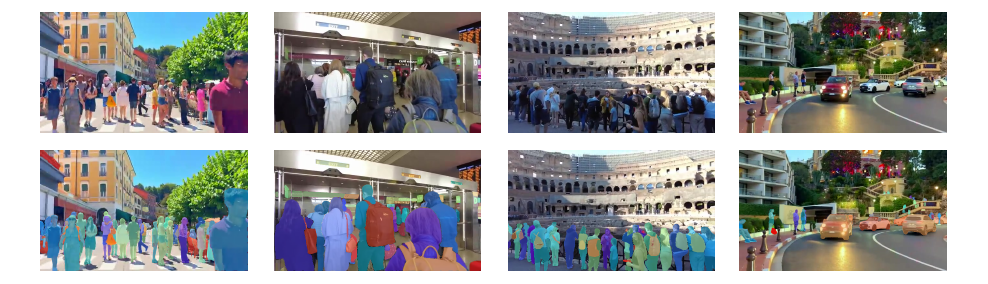} %
    \vspace{-2.5em}
  \caption{\textbf{EpicTours dataset}: Our dataset consists of up to $3$ hour long videos of walking tours across cities spanning the globe. We provide manual annotations at semantic/instance level for studying TTT on videos. Dataset shall be made publicly available. }
  \vspace{-2em}
  \label{fig:dataset}
\end{figure}

\noindent To address this, we propose a new video segmentation dataset, \textbf{EpicTours Dataset}, which contains multi-hour-long videos of people exploring cities across the globe, representing geographical diversity. Our videos are densely annotated with the help of SAM 3 inference \cite{carion2025sam}, manually filtered/refined by expert human annotators. Notably, our EpicTours dataset contains videos up to 3 hours long, with $30$ classes (subset of COCO). This enables us to study the applicability of our FFN on real-world scenarios, for example, videos containing multiple-people. Please refer to the supplementary for more details. \\

\begin{table}[t]
\centering
\caption{\textbf{TTT Results for segmentation.} Time is in seconds per frame (A100 GPU). s.w: sliding window holding $k$ frames for TTT. no s.w.: TTT is done non-overlapping sliding windows. FFN offers competitive performance with lower time. $^\dagger$: effects of individual components analyzed later.
}
\label{tab:ttt_segmentation}
\vspace{-0.5em}
\resizebox{0.7\columnwidth}{!}{
\begin{tabular}{l l c c c c c}
\toprule
\multirow{2}{*}{Setting} & \multirow{2}{*}{Method} & \multicolumn{2}{c}{COCO Videos} & \multicolumn{3}{c}{KITTI-STEP} \\
\cmidrule(lr){3-4} \cmidrule(lr){5-7}
& & \emph{Inst.$\uparrow$} & \emph{Pan.$\uparrow$} & \emph{Val.$\uparrow$} & \emph{Test$\uparrow$} & \emph{Time$\downarrow$} \\
\midrule
\multirow{3}{*}{Independent frames} & Main Task Only & 16.7 & 13.9 & 53.8 & 52.5 & 1.8 \\
& MAE Joint Training & 16.5 & 13.5 & 53.5 & 52.5 & 1.8 \\
& TTT-MAE No Memory & 35.4 & 20.1 & 53.6 & 52.5 & 3.8 \\
\midrule
Full Video & Offline TTT-MAE All Frames & 33.6 & 19.6 & 53.2 & 51.2 & 1.8 \\
\midrule
\multirow{6}{*}{Stream} & LN Adapt & 16.5 & 14.7 & 53.8 & 52.5 & 2.0 \\  
& Tent & 16.6 & 14.6 & 53.8 & 52.2 & 2.8 \\  
& Tent w/ Class Bal. & 16.7 & 14.8 & 53.8 & 52.5 & 3.7 \\
& Self-Train & - & - & 54.7 & 54.0 & 6.6 \\ 
& Self-Train w/ Class Bal. & - & - & 54.1 & 53.6 & 6.9 \\
& Online TTT-MAE (n.o. s.w.) & 35.3  & 20.8 & 48.1 & 51.7 & 0.4 \\
& Online TTT-MAE (s.w.) & 37.6 & 21.7 & 55.4 & 54.3 & 4.1 \\
\cmidrule{2-7}
\rowcolor{gray!15}
& FFN (Ours)$^\dagger$ & 45.1 & 29.6 & 57.3 & 59.5 & 0.7\\
\bottomrule
\end{tabular}
}
\vspace{-2em}
\end{table}

\noindent \textbf{Metrics}: For instance, panoptic and semantic segmentation we report Average Precision (AP), Panoptic Quality (PQ), and mIoU respectively. Similarly, for video-depth estimation, we report AbsRel:Absolute Relative Error, and $\delta_1$ which denotes $\%$ of pixels where the ratio between the prediction and the ground truth falls below a threshold of $1.25$. For action-classification, we report top-1 accuracy. \\

\begin{wraptable}{r}{0.35\textwidth} %
    \centering
    \scriptsize
    \vspace{-2.5em} %
    \caption{\textbf{EpicTours Dataset.} Our dataset contains videos averaging 1.5 hours with dense annotations. len means average length in seconds.}
    \label{tab:dataset_details}
    \begin{tabular}{@{}lcc@{}}
        \toprule
        Video Dataset & Len (s) $\uparrow$& Frames$\uparrow$ \\
        \midrule
        CityScapes~\cite{7780719} & 1.8 & 3k \\
        DAVIS~\cite{Perazzi2016davis} & 3.5 & 3.4k \\
        YouTube~\cite{xu2018youtube} & 4.5 & 123k \\
        KITTI-S~\cite{kong2023robodepth} & 40.0 & 8k \\
        COCO-V                    & 309.0 & 30k \\
        \midrule
        \rowcolor{gray!15}
        EpicTours & 5300.4 & 2.49M \\
        \bottomrule
    \end{tabular}
    \vspace{-2.5em} %
\end{wraptable}

\noindent \textbf{Implementation details:} For segmentation (Tab\ref{tab:ttt_segmentation}, Tab\ref{tab:epic_tour_results}), we initialize the backbone $f$ with Mask2former, SSL-task head $g$ from scratch. We train $g$ jointly during the training-time phase on both image datasets / a single long video. Similarly, $h$ the task-specific head (segmentation) utilizes Mask2former's pre-trained weights. For depth-estimation (Tab\ref{tab:depth_estimation}), we leverage the Video-Depth Anything architecture. For video-classification (Tab\ref{tab:classification}), we adopt the self-supervised backbone (DINOv3)\cite{simeoni2025dinov3} backbone (a very-recent foundational model). We set the memory buffer size $W=50$, perform a \textit{single} gradient step per incoming-frame, and use the Adam optimizer. Our SSL-head $g$ is trained with next-frame anticipation instead of current-frame reconstruction. We perform experiments with $3$ seeds and report their mean accuracy. All experiments are performed on a single ampere of 80 GB. \\

\vspace{-1em}
\noindent \textbf{Baselines:} Inspired by \cite{wang2025test}, we consider multiple baselines that \textit{span} image-models, offline video processing, test-time-adaptation (TTA), test-time-training. In \ref{tab:ttt_segmentation}, we categorize our baselines as (i) independent frames: the main task only means zero-shot per-frame inference on a model trained on image-segmentation. MAE joint-training \textit{jointly} pretrains MAE\cite{mae} for reconstruction/segmentation. Similarly, TTT-MAE memory takes the \textit{previous} MAE-joint training baseline, performs TTT on \textit{each} frame \textit{independently} (ii) full-video: an ideal \textit{offline-oracle}, allowed to access the \textit{entire} video (iii) stream: a tougher setup where a model only encounters frames one-by-one, `without' peeking in the future.\\

\noindent Our TTA baselines include the pioneering TENT approach\cite{wang2020tent}, layer-norm adaptation, self-training with confident pseudo-labels (to rule out whether improvements are due to self-supervision or TTT on videos). Next, we consider the TTT-Online baseline \cite{wang2025test} across two setups (a) without sliding window: we perform TTT on a window of $w$ frames and then move onto a \textit{non-overlapping} window; (b) with overlapping sliding window: the original setup in \cite{wang2025test}. Finally, we consider zero-shot \textit{foundational} models like DinoV3, evaluating feature robustness on general representation-learning.\\  

\noindent \textbf{FFN can surpass TTT-online on segmentation:} In Tab \ref{tab:ttt_segmentation} for the COCO-Videos dataset, our FFN obtains $45.1 (+7.5\uparrow)$ for instance segmentation, $29.6$ $(+7.9\uparrow)$ for panoptic segmentation. Similarly, on the KITTI-STEP validation set, FFN gets $57.3 (+1.9\uparrow)$, $59.5(+5.2\uparrow)$ on the testing split. One takeaway is that the \textit{streaming} baselines (including our FFN) can \textit{surpass} the offline TTT-MAE oracle, which \textit{can access} entire video, thereby validating the principle-of-locality. Similarly, FFN performs competitively with TTA baselines like Tent.\\

\begin{table*}[t]
\centering
\caption{\textbf{Generalization of TTT to video-depth estimation.} Table shows zero-shot baselines containing both single image~\cite{depth_anything_v2} and video depth estimation models~\cite{wang2023neural, chronodepth, hu2024depthcrafter, depthanyvideo}. For TTT baselines, we benchmark the strongest baseline (TTT-Online). Results may indicate FFN's competitiveness with current state of the art. (s.w.): denotes sliding window. ($^\dagger$): contains 50 frames. ($^\ddagger$): contains 170 frames. \cite{depth_anything_v2,hu2024depthcrafter}. }
\label{tab::quant_video_depth_benchmark}
\resizebox{\textwidth}{!}{
\begin{tabular}{l c c c c c c c c c c c}
\toprule
\multirow{2}{*}{Method / Metrics} & \multicolumn{2}{c}{KITTI~\cite{geiger2013vision}} & \multicolumn{2}{c}{Scannet~\cite{dai2017scannet}} & \multicolumn{2}{c}{Bonn~\cite{palazzolo2019iros}} & \multicolumn{2}{c}{NYUv2~\cite{Silberman:ECCV12}} & \multicolumn{2}{c}{Sintel~\cite{Butler_Wulff_Stanley_Black_2012} $^\dagger$} & \multicolumn{1}{c}{Scannet $^\ddagger$} \\
\cmidrule(lr){2-3} \cmidrule(lr){4-5} \cmidrule(lr){6-7} \cmidrule(lr){8-9} \cmidrule(lr){10-11} \cmidrule(lr){12-12}
& AbsRel~(↓) & $\delta_1$~(↑) & AbsRel~(↓) & $\delta_1$~(↑) & AbsRel~(↓) & $\delta_1$~(↑) & AbsRel~(↓) & $\delta_1$~(↑) & AbsRel~(↓) & $\delta_1$~(↑) & TAE~(↓) \\
\midrule 
\textit{Zero-shot baselines}\\
DAv2-L~\cite{depth_anything_v2} & 0.137 & 0.815 & 0.150 & 0.768 & 0.127 & 0.864 & 0.094 & 0.928 & 0.390 & 0.541 & 1.140 \\
NVDS~\cite{wang2023neural} & 0.233 & 0.614 & 0.207 & 0.628 & 0.199 & 0.674 & 0.217 & 0.598 & 0.408 & 0.464 & 2.176 \\
NVDS + DAv2-L & 0.227 & 0.617 & 0.194 & 0.658 & 0.191 & 0.700 & 0.184 & 0.679 & 0.449 & 0.503 & 2.536 \\
ChronoDepth~\cite{chronodepth} & 0.243 & 0.576 & 0.199 & 0.665 & 0.199 & 0.665 & 0.173 & 0.771 & 0.192 & 0.673 & 1.022 \\
DepthCrafter~\cite{hu2024depthcrafter} & 0.164 & 0.753 & 0.169 & 0.730 & 0.153 & 0.803 & 0.141 & 0.822 & 0.299 & 0.695 & 0.639 \\
DepthAnyVideo~\cite{depthanyvideo} & - & - & - & - & - & - & - & - & 0.405 & 0.659 & 0.967 \\
Video Depth-Anything~\cite{chen2025video} & 0.083 & 0.946 & 0.087 & 0.933 & 0.070 & 0.961 & 0.064 & 0.967 & 0.300 & 0.633 & 0.570\\
\midrule 
\textit{TTT baselines}\\
Online TTT-MAE (s.w.)& 0.071 & 0.958 & 0.076 & 0.949 & 0.064 & 0.970 & 0.058 & 0.974 & 0.265 & 0.710 & 0.495 \\
\rowcolor{gray!15}
FFN (Ours)  & 0.059 & 0.972 & 0.062 & 0.965 & 0.051 & 0.982 & 0.049 & 0.988 & 0.230 & 0.762 & 0.380 \\
\bottomrule
\end{tabular}
\label{tab:depth_estimation}
}
\end{table*}

\begin{table}[t]
    \centering
    \begin{minipage}[t]{0.5\textwidth}
        \centering
        \caption{\textbf{TTT results for coarse-classification }on video-action datasets. We report top 1 accuracy. }
        \label{tab:classification}
        \scriptsize
        \setlength{\tabcolsep}{10pt} %
        \vspace{-1em}
        \begin{tabular}{@{}lcc@{}}
            \toprule
            Method & UCF101$\uparrow$ & SSv2$\uparrow$ \\
            \midrule
            \textit{TTA baselines}\\
            DINOv2   & 93.8 & 68.4 \\
            V-JEPA 2 & 93.8 & 75.4 \\
            Web-DINO & 94.1 & 68.1 \\
            DINOv3   & 93.5 & 70.8 \\
            \midrule
            \textit{TTT-baselines}\\
            Online TTT & 94.2 & 73.4 \\
            \midrule
            \rowcolor{gray!15}
            FFN (Ours) & 95.0 & 74.1 \\
            \bottomrule
        \end{tabular}
    \end{minipage}
    \hfill
    \begin{minipage}[t]{0.45\textwidth}
        \centering
        \caption{\textbf{TTT results on hours long videos}: evaluated on proposed EpicTours dataset.}
        \label{tab:epic_tour_results}
        \scriptsize
        \setlength{\tabcolsep}{10pt} %
        \vspace{-0.75em}
        \begin{tabular}{@{}lcc@{}}
            \toprule
            Method & Sem.$\uparrow$ & Inst.$\uparrow$ \\
            \midrule
            \textit{TTA/TTT baselines}\\
            LN-Adapt & 32.1 & 26.8 \\
            Tent & 35.6 & 27.7 \\
            Tent w/ Bal. & 35.9 &29.3\\
            Self-Train & 37.3 & -- \\
            Self-Train w/ Bal. & 39.1 & -- \\
            TTT-Online (s.w.) & 42.8 & 31.4 \\
            \midrule
            \rowcolor{gray!15}
            FFN (Ours) & 49.3 & 36.7 \\
            \bottomrule
        \end{tabular}
    \end{minipage}
    \vspace{-1em}
\end{table}

\noindent \textbf{FFN for performing TTT with better compute-accuracy tradeoffs:} One concern about TTT is that it involves backpropagation through both backbone $f$, SSL-head $g$, for \textit{every} incoming frame, making it \textit{slow} in practice. For example, as shown in Tab\ref{tab:ttt_segmentation}, TTT-Online takes $4.1sec$ \textit{per} frame. Fortunately, FFN processes only $3$ frames per window and consumes $0.7sec$ per timestep. It also gets a higher result ($45.1$ vs $37.6$) showing a \textit{better} compute-accuracy tradeoff. We acknowledge the potential to make TTT \textit{even} faster.\\

\noindent \textbf{FFN can generalize to video depth-estimation:} We test the generalization ability of FFN on a challenging video depth-estimation task (Table \ref{tab:depth_estimation}). In particular, FFN performs well across the board on all $6$ datasets.\\

\noindent \textbf{FFN can classify coarse-actions in videos: } While segmentation only tests per-frame fine-grained understanding, we also study how well FFN understands the \textit{temporal} aspect. Thus, we subject FFN to action-classification in UCF101/ Something-Somethingv2. We notice gains of $0.8\uparrow$, $0.7\uparrow$, respectively. Note that SSv2 is especially challenging; it requires FFN to reason about \textit{direction} of time: e.g., moving something up vs down.\\

 \noindent \textbf{FFN's effectiveness on \textit{realistic} \textit{hours} long videos:} In Tab\ref{tab:epic_tour_results}, we subject our FFN to an even tougher task: how well can it adapt to hours long videos captured in the real-world, often on \textit{low-resolution} devices like cellphones. In the EpicTours dataset, FFN shows promising gains of $6.5\%\uparrow$ for semantic segmentation, $5.3\%\uparrow$ for instance segmentation. A use-case \textit{may be} to deploy FFN on drones for disaster-relief efforts or food-delivery.

\section{Ablations on FFN}
Here, we discuss ablations on the Frame Forgetting Network presented in Tab\ref{tab:ttt_segmentation}.\\

\noindent \textbf{Rope encoding performs best:} Recall that FFN injects temporal information about an incoming frame by conditioning the backbone $f$ with time (Fig \ref{fig:teaser_wrap}). Table \ref{tab:ablation_pe} shows that the the rope-based time encoding outperforms both relative and absolute time encodings. Intuitively, rope can rotate both query/key matrices in latent space, and encode both relative/absolute positions \textit{simultaneously}.\\

\begin{table*}[t!]
    \centering
    \begin{minipage}{0.31\textwidth}
        \centering
        \caption{Impact of different choices of positional encodings.}
        \label{tab:ablation_pe}
        \vspace{-0.5em}
        \resizebox{\textwidth}{!}{
            \begin{tabular}{l cc cc}
                \toprule
                \multirow{2}{*}{Method} & \multicolumn{2}{c}{COCO} & \multicolumn{2}{c}{KITTI} \\
                \cmidrule(lr){2-3} \cmidrule(lr){4-5}
                & \emph{Inst.$\uparrow$} & \emph{Pan.$\uparrow$} & \emph{Val.$\uparrow$} & \emph{Test$\uparrow$} \\
                \midrule
                + relative & 40.5 & 24.1 & 53.9 & 54.8 \\
                + absolute$^\dagger$ & 45.1 & 29.6 & 57.3 & 59.5 \\
                \rowcolor{gray!15}
                + rope     & 46.8 & 31.2 & 58.9 & 60.7 \\
                \bottomrule
                \vspace{-0.5em}
            \end{tabular}
        }
    \end{minipage}
    \hfill
    \begin{minipage}{0.31\textwidth}
        \centering
        \caption{ Impact of various choices on  memory buffer $F$.}
        \label{tab:ablation_memory_buffer}
        \vspace{-0.5em}
        \resizebox{\textwidth}{!}{
            \begin{tabular}{l cc cc}
                \toprule
                \multirow{2}{*}{Memory Buffer $F$} & \multicolumn{2}{c}{COCO} & \multicolumn{2}{c}{KITTI} \\
                \cmidrule(lr){2-3} \cmidrule(lr){4-5}
                & \emph{Inst.$\uparrow$} & \emph{Pan.$\uparrow$} & \emph{Val.$\uparrow$} & \emph{Test$\uparrow$} \\
                \midrule
                FIFO      & 43.9 & 28.1 & 56.4 & 57.8 \\
                MBO$^\dagger$     & 45.1 & 29.6 & 57.3 & 59.5 \\
                \rowcolor{gray!15}
                FIFO+MBO & 45.7 & 31.9 & 59.1 & 61.0 \\
                \bottomrule
            \end{tabular}
        }
    \end{minipage}
    \hfill
    \begin{minipage}{0.31\textwidth}
        \centering
        \caption{Impact of different losses for computing $f_{latent}$.}
        \label{tab:latent_loss}
        \vspace{-0.5em}
        \resizebox{\textwidth}{!}{
            \begin{tabular}{l cc cc}
                \toprule
                \multirow{2}{*}{Method} & \multicolumn{2}{c}{COCO} & \multicolumn{2}{c}{KITTI} \\
                \cmidrule(lr){2-3} \cmidrule(lr){4-5}
                & \emph{Inst.$\uparrow$} & \emph{Pan.$\uparrow$} & \emph{Val.$\uparrow$} & \emph{Test$\uparrow$} \\
                \midrule
                + $L_1$   & 43.2 & 27.8 & 56.1 & 57.4 \\
                + $L_2$   & 44.5 & 28.9 & 56.8 & 58.6 \\
                \rowcolor{gray!15}
                + Cosine$^\dagger$  & 45.1 & 29.6 & 57.3 & 59.5 \\
                \bottomrule
            \end{tabular}
        }
    \end{minipage}
\end{table*}

\begin{figure}[h!] %
  \centering
  \vspace{-2em}
  \includegraphics[width=\textwidth]{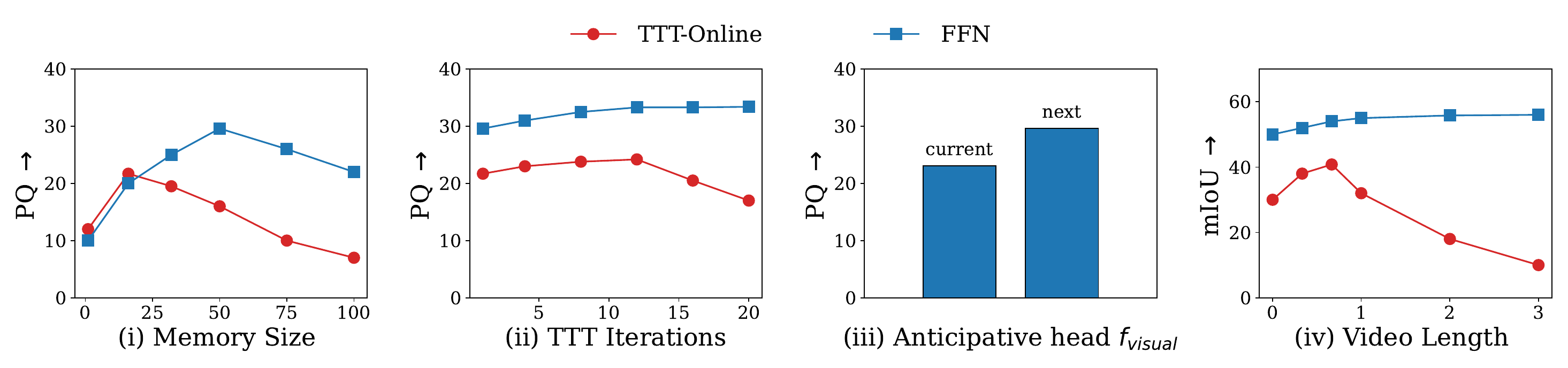} %
  \caption{First three plots show panoptic segmentation on COCO-Videos whereas last plot shows semantic segmentation on our EpicTours dataset. (i) Effect of increasing the size of buffer $B$ (ii) Increasing number of iterations on each test sample during TTT (iii) Effect of training SSL head with current-frame reconstruction/ vs next-frame. (iv) FFN's performance remains stable even when subjected to $3$ hour long videos, whereas TTT-Online degrades rapidly.
  }
  \vspace{-2.5em}
  \label{fig:ablations}
\end{figure}

\noindent \textbf{Temporal conditioning of the backbone $f$ helps:} We perform an ablation where we remove $t$ as a condition of the backbone $f$ and find that performance \textit{drops} from 45.1 to $42.7$ on COCO-Videos instance segmentation. This shows that instilling a notion of time in $f$ helps.\\

 \noindent \textbf{Memory buffer B can take inspiration from both FIFO/MBC policies:} In Table \ref{tab:ablation_memory_buffer}, we implement two variations of the memory buffer $B$. MBC means that an incoming frame is merged with the `most-similar' frame. MBC performs better than the FIFO queue\cite{he2024ma}. Alternatively, one may also `merge' an incoming frame (MBC) and consider it as `most recent' element in a FIFO-queue\footnote{After merging, we swap the merged position with \textit{last position} in the FIFO queue (similar to the quicksort algorithm). This prevents the bottleneck of shifting \textit{all elements} one by one to achieve sorting. The former operation is a single swap, whereas the latter is computationally expensive (and CPU becomes the bottleneck instead of GPU).}, achieving even \textit{better} results. \\

  \noindent \textbf{Cosine loss performs best on $v_{latent}$}: In Tab\ref{tab:latent_loss}, we experiment with different types of loss to compute $v_{latent}$ (in eq\ref{eq:f_latent}), and find that cosine works the best. Intuitively, cosine projects all features to a unit-hypersphere, cancels out their magnitudes, and only measures angle difference, leading to a better estimate.\\

 \noindent \textbf{Increasing buffer length helps, but \textit{only} up to a limit:} In Fig\ref{fig:ablations}(i), we see that FFN performs well as buffer-size increases up to $50$ frames, then drops. This helps validate the insight that keeping only \textit{some} past-frames is important, and keeping everything \textit{worsens} the performance. Intuitively, the capacity of the model $f$ is \textit{finite}, trying to remember everything leads to instability during learning. \\

 \noindent \textbf{Increasing TTT iterations improves performance:} Fig\ref{fig:ablations}(ii) shows that increasing TTT iterations improves performance. Note that FFN remains \textit{almost constant}, which means that it manages to learn a lot even with just one gradient-step, whereas TTT-online needs several steps to reach its peak.\\

 \noindent \textbf{Anticipating the `next' frame is better than `current' frame prediction in $f_{visual}$}: Fig\ref{fig:ablations}(iii) shows that it is better to predict what shall come one step in the future, rather than just predicting the current frame, thereby showing that `anticipation' can provide a unique inductive bias to a neural net.\\

 \noindent \textbf{FFN retains stable performance for long videos:} In Fig\ref{fig:ablations}(iv), we plot FFN's average performance as it continues to perform TTT over 3 hour long videos on our EpicTour dataset. Note that TTT-online degrades beyond 50 mins, whereas FFN improves/retains performance over time, indicating drift is less of an issue as compared to TTT-online. We refer the reader to supplementary material for additional ablations. \\

\section{Related Work}

\noindent  There are several works operating within the regime of `online model distillation' \cite{mullapudi2018online}: here the teacher is generally kept on the Internet, while a student sitting on an edge-device makes \textit{decision} when to adapt or not. In FFN, we aim to adapt a \textit{single model} on-device and don't require a teacher. Moreover, we focus on improving the quality of generated predictions and not explicitly matching the real-time performance of detectors, which remains an open challenge\cite{wang2025test}. \\

\noindent The idea of `principle of locality' can be traced back to Vapnik et al \cite{Gammerman98learningby, vapnik_book}. This also found applications in \cite{bottou1992local,zhang2006svm,hardt2023test,bottou1992local}. FFN's difference lies in how it \textit{implements} this locality: methods like TTT-Online\cite{wang2025test} `reset' weights after certain set of frames is processed. However, FFN's rely on the `forgetting' principle: restore a model's baseline prior to adaptation. Although both aim to achieve identical goals (locality), the FFN's mechanism is different. While ideas on forgetting are typically used to unlearn `bad' representations in safety-alignment\cite{qi2024safety}, we showcase the application of this principle for \textit{representational-restoration}.\\

\noindent Similarly, the computer vision community has used the idea of test-time-training for several applications ~\cite{jain2011online, shocher2018zero, nitzan2022mystyle, xie2023sepico}, 
especially depth estimation~\cite{tonioni2019learning, tonioni2019real, zhang2020online, zhong2018open, luo2020consistent}. Other works such as \cite{hansen2020self, sun2021online, liu2021ttt++, yuan2023robust} explore online-learning and their extensions to videos \cite{volpi2022road}. Although the videos considered in such papers are mostly synthetic/very short, we also show results on long-realistic videos. \\

\noindent Finally, video compression algorithms \cite{wiegand2003overview} also implement a mechanism to measure surprise: they measure how much change a new frame offers relative to a previous frame, which in turn is used to inform compression. The key idea is that most redundant frames are assigned short length codes\cite{sutton1995generalization}. However, this process is typically \textit{not} learned, whereas FFN \textit{learns} via an anticipative-head. Furthermore, in video-compression, the size of the file \textit{grows} with the number of frames encountered, whereas in FFN all the knowledge of a video is squeezed into a \textit{finite} set of weights\cite{hinton2022forward}. 

\section{Additional Discussions}
We  shall now discuss some directions that might help \textit{inspire} further research on FFN. Please note that while these are relevant to FFN, their precise implementation remains beyond the scope of this work.
\noindent Although TTT requires only \textit{one} gradient step, it \textit{remains slow} in practice. One reason is that it still relies on backpropagation. It creates a subtle `layer-lock'\cite{lowe2019putting}: the first layer has to `wait' for the gradients from the last layer to come back, thereby wasting compute cycles. Alternatives include local-learning algorithms like forward-forward\cite{hinton2022forward}, target propagation\cite{lee2015difference} or no-prop \cite{li2025noprop}, which can update these weights during the feed-forward phase only. A key challenge remains that these algorithms still do not surpass backpropagation's performance on large scale datasets.\\

\noindent TTT relies on the assumption that the `useful' video frames keep `streaming' continuously. However, there might be cases where there are \textit{no changes} in the incoming frames (e.g., stationary cameras). During that time, learning does not happen, and FFN remains \textit{idle}. An alternate mechanism was discussed by \cite{hinton2021glomgodfather,hinton1995wake}: make FFN enter a "sleeping" phase, sample the data from within FFN itself, and perform a \textit{few} iterations of gradient descent. A key challenge remains how to "sample" efficiently\cite{hinton1984boltzmann}, which was later \textit{partially} resolved by score-matching networks relying on Langevin dynamics\cite{song2020score}. 
\noindent Further, we showed that anticipating the next-frame works \textit{better} than the current-frame prediction. This lends itself to the question: Is it \textit{possible} to meta-learn the SSL task itself. Recent works like TTT-MLP have just started exploring this interesting direction\cite{sun2024learning}.

\section{Conclusion}
In this work, we introduced the Frame Forgetting Network (FFN): for performing test-time-training on videos which may be hours long. That is, a memory restoration mechanism allows the FFN to restore the representation of a model before adaptation. Our key contribution is the logic of handling sliding temporal-windows by operating selectively on exiting and entering frames, rather than doing duplicate processing at each time-step.\\

\noindent Next, we discuss an adaptive anticipative head that decides when to do TTT on an incoming frame. Finally, we introduced the EpicTours dataset, which contains \textit{hours long videos} to effectively study this important problem. Our empirical results and ablations may help validate further merits of such an approach. In the future, we hope to explore FFN for hours long video generation \cite{dalal2025one}.\\

\bibliographystyle{unsrt}
\bibliography{main}

\begin{thebibliography}{10}

\bibitem{oquab2023dinov2}
Maxime Oquab, Timoth{\'e}e Darcet, Th{\'e}o Moutakanni, Huy Vo, Marc
  Szafraniec, Vasil Khalidov, Pierre Fernandez, Daniel Haziza, Francisco Massa,
  Alaaeldin El-Nouby, et~al.
\newblock Dinov2: Learning robust visual features without supervision.
\newblock {\em arXiv preprint arXiv:2304.07193}, 2023.

\bibitem{caron2021dino}
Mathilde Caron, Hugo Touvron, Ishan Misra, Herv{\'e} J{\'e}gou, Julien Mairal,
  Piotr Bojanowski, and Armand Joulin.
\newblock {Emerging Properties in Self-Supervised Vision Transformers}.
\newblock In {\em ICCV}, 2021.

\bibitem{mullapudi2018online}
Ravi~Teja Mullapudi, Steven Chen, Keyi Zhang, Deva Ramanan, and Kayvon
  Fatahalian.
\newblock Online model distillation for efficient video inference.
\newblock {\em arXiv preprint arXiv:1812.02699}, 2018.

\bibitem{wang2025test}
Renhao Wang, Yu~Sun, Arnuv Tandon, Yossi Gandelsman, Xinlei Chen, Alexei~A
  Efros, and Xiaolong Wang.
\newblock Test-time training on video streams.
\newblock {\em Journal of Machine Learning Research}, 26(9):1--29, 2025.

\bibitem{bentley1984programming}
Jon Bentley.
\newblock Programming pearls: algorithm design techniques.
\newblock {\em Communications of the ACM}, 27(9):865--873, 1984.

\bibitem{hinton1986learning}
Geoffrey~E Hinton.
\newblock Learning distributed representations of concepts.
\newblock In {\em Proceedings of the Annual Meeting of the Cognitive Science
  Society}, volume~8, 1986.

\bibitem{mildenhall2021nerf}
Ben Mildenhall, Pratul~P Srinivasan, Matthew Tancik, Jonathan~T Barron, Ravi
  Ramamoorthi, and Ren Ng.
\newblock Nerf: Representing scenes as neural radiance fields for view
  synthesis.
\newblock {\em Communications of the ACM}, 65(1):99--106, 2021.

\bibitem{vaswani2017attention}
Ashish Vaswani, Noam Shazeer, Niki Parmar, Jakob Uszkoreit, Llion Jones,
  Aidan~N Gomez, {\L}ukasz Kaiser, and Illia Polosukhin.
\newblock Attention is all you need.
\newblock {\em Advances in neural information processing systems}, 30, 2017.

\bibitem{zhang2019lookahead}
Michael Zhang, James Lucas, Jimmy Ba, and Geoffrey~E Hinton.
\newblock Lookahead optimizer: k steps forward, 1 step back.
\newblock {\em Advances in neural information processing systems}, 32, 2019.

\bibitem{oord2018representation}
Aaron van~den Oord, Yazhe Li, and Oriol Vinyals.
\newblock Representation learning with contrastive predictive coding.
\newblock {\em arXiv preprint arXiv:1807.03748}, 2018.

\bibitem{wei2018learning}
Donglai Wei, Joseph~J Lim, Andrew Zisserman, and William~T Freeman.
\newblock Learning and using the arrow of time.
\newblock In {\em Proceedings of the IEEE conference on computer vision and
  pattern recognition}, pages 8052--8060, 2018.

\bibitem{layzer1975arrow}
David Layzer.
\newblock The arrow of time.
\newblock {\em Scientific American}, 233(6):56--69, 1975.

\bibitem{wei2017collaborative}
Jian Wei, Jianhua He, Kai Chen, Yi~Zhou, and Zuoyin Tang.
\newblock Collaborative filtering and deep learning based recommendation system
  for cold start items.
\newblock {\em Expert systems with applications}, 69:29--39, 2017.

\bibitem{weber2021step}
Mark Weber, Jun Xie, Maxwell Collins, Yukun Zhu, Paul Voigtlaender, Hartwig
  Adam, Bradley Green, Andreas Geiger, Bastian Leibe, Daniel Cremers, et~al.
\newblock Step: Segmenting and tracking every pixel.
\newblock {\em arXiv preprint arXiv:2102.11859}, 2021.

\bibitem{ucf101}
Khurram Soomro, Amir~Roshan Zamir, and Mubarak Shah.
\newblock Ucf101.
\newblock In {\em arXiv preprint arXiv:1212.0402}, 2012.

\bibitem{goyal2017something}
Raghav Goyal, Samira Ebrahimi~Kahou, Vincent Michalski, Joanna Materzynska,
  Susanne Westphal, Heuna Kim, Valentin Haenel, Ingo Fruend, Peter Yianilos,
  Moritz Mueller-Freitag, et~al.
\newblock The" something something" video database for learning and evaluating
  visual common sense.
\newblock In {\em Proceedings of the IEEE international conference on computer
  vision}, pages 5842--5850, 2017.

\bibitem{yang2024depth}
Lihe Yang, Bingyi Kang, Zilong Huang, Zhen Zhao, Xiaogang Xu, Jiashi Feng, and
  Hengshuang Zhao.
\newblock Depth anything v2.
\newblock {\em Advances in Neural Information Processing Systems},
  37:21875--21911, 2024.

\bibitem{geiger2013vision}
Andreas Geiger, Philip Lenz, Christoph Stiller, and Raquel Urtasun.
\newblock Vision meets robotics: The kitti dataset.
\newblock {\em The International Journal of Robotics Research},
  32(11):1231--1237, 2013.

\bibitem{dai2017scannet}
Angela Dai, Angel~X Chang, Manolis Savva, Maciej Halber, Thomas Funkhouser, and
  Matthias Nie{\ss}ner.
\newblock Scannet: Richly-annotated 3d reconstructions of indoor scenes.
\newblock In {\em Proceedings of the IEEE conference on computer vision and
  pattern recognition}, pages 5828--5839, 2017.

\bibitem{palazzolo2019iros}
E.~Palazzolo, J.~Behley, P.~Lottes, P.~Gigu\`ere, and C.~Stachniss.
\newblock {ReFusion: 3D Reconstruction in Dynamic Environments for RGB-D
  Cameras Exploiting Residuals}.
\newblock 2019.

\bibitem{Silberman:ECCV12}
Pushmeet~Kohli Nathan~Silberman, Derek~Hoiem and Rob Fergus.
\newblock Indoor segmentation and support inference from rgbd images.
\newblock In {\em ECCV}, 2012.

\bibitem{Butler_Wulff_Stanley_Black_2012}
Daniel~J. Butler, Jonas Wulff, Garrett~B. Stanley, and Michael~J. Black.
\newblock {\em A Naturalistic Open Source Movie for Optical Flow Evaluation},
  page 611–625.
\newblock Jan 2012.

\bibitem{carion2025sam}
Nicolas Carion, Laura Gustafson, Yuan-Ting Hu, Shoubhik Debnath, Ronghang Hu,
  Didac Suris, Chaitanya Ryali, Kalyan~Vasudev Alwala, Haitham Khedr, Andrew
  Huang, et~al.
\newblock Sam 3: Segment anything with concepts.
\newblock {\em arXiv preprint arXiv:2511.16719}, 2025.

\bibitem{7780719}
Marius Cordts, Mohamed Omran, Sebastian Ramos, Timo Rehfeld, Markus Enzweiler,
  Rodrigo Benenson, Uwe Franke, Stefan Roth, and Bernt Schiele.
\newblock The cityscapes dataset for semantic urban scene understanding.
\newblock In {\em 2016 IEEE Conference on Computer Vision and Pattern
  Recognition (CVPR)}, pages 3213--3223, 2016.

\bibitem{Perazzi2016davis}
F.~Perazzi, J.~Pont-Tuset, B.~McWilliams, L.~{Van Gool}, M.~Gross, and
  A.~Sorkine-Hornung.
\newblock A benchmark dataset and evaluation methodology for video object
  segmentation.
\newblock In {\em Computer Vision and Pattern Recognition}, 2016.

\bibitem{xu2018youtube}
Ning Xu, Linjie Yang, Yuchen Fan, Jianchao Yang, Dingcheng Yue, Yuchen Liang,
  Brian Price, Scott Cohen, and Thomas Huang.
\newblock Youtube-vos: Sequence-to-sequence video object segmentation.
\newblock In {\em Proceedings of the European conference on computer vision
  (ECCV)}, pages 585--601, 2018.

\bibitem{kong2023robodepth}
Lingdong Kong, Shaoyuan Xie, Hanjiang Hu, Lai~Xing Ng, Benoit Cottereau, and
  Wei~Tsang Ooi.
\newblock Robodepth: Robust out-of-distribution depth estimation under
  corruptions.
\newblock {\em Advances in Neural Information Processing Systems},
  36:21298--21342, 2023.

\bibitem{simeoni2025dinov3}
Oriane Sim{\'e}oni, Huy~V Vo, Maximilian Seitzer, Federico Baldassarre, Maxime
  Oquab, Cijo Jose, Vasil Khalidov, Marc Szafraniec, Seungeun Yi, Micha{\"e}l
  Ramamonjisoa, et~al.
\newblock Dinov3.
\newblock {\em arXiv preprint arXiv:2508.10104}, 2025.

\bibitem{mae}
Kaiming He, Xinlei Chen, Saining Xie, Yanghao Li, Piotr Doll{\'{a}}r, and
  Ross~B. Girshick.
\newblock Masked autoencoders are scalable vision learners.
\newblock {\em CoRR}, abs/2111.06377, 2021.

\bibitem{wang2020tent}
Dequan Wang, Evan Shelhamer, Shaoteng Liu, Bruno Olshausen, and Trevor Darrell.
\newblock Tent: Fully test-time adaptation by entropy minimization.
\newblock {\em arXiv preprint arXiv:2006.10726}, 2020.

\bibitem{depth_anything_v2}
Lihe Yang, Bingyi Kang, Zilong Huang, Zhen Zhao, Xiaogang Xu, Jiashi Feng, and
  Hengshuang Zhao.
\newblock Depth anything v2.
\newblock {\em arXiv:2406.09414}, 2024.

\bibitem{wang2023neural}
Yiran Wang, Min Shi, Jiaqi Li, Zihao Huang, Zhiguo Cao, Jianming Zhang,
  Ke~Xian, and Guosheng Lin.
\newblock Neural video depth stabilizer.
\newblock In {\em Proceedings of the IEEE/CVF International Conference on
  Computer Vision}, pages 9466--9476, 2023.

\bibitem{chronodepth}
Jiahao Shao, Yuanbo Yang, Hongyu Zhou, Youmin Zhang, Yujun Shen, Matteo Poggi,
  and Yiyi Liao.
\newblock Learning temporally consistent video depth from video diffusion
  priors.
\newblock {\em arXiv preprint arXiv:2406.01493}, 2024.

\bibitem{hu2024depthcrafter}
Wenbo Hu, Xiangjun Gao, Xiaoyu Li, Sijie Zhao, Xiaodong Cun, Yong Zhang, Long
  Quan, and Ying Shan.
\newblock Depthcrafter: Generating consistent long depth sequences for
  open-world videos.
\newblock {\em arXiv preprint arXiv:2409.02095}, 2024.

\bibitem{depthanyvideo}
Honghui Yang, Di~Huang, Wei Yin, Chunhua Shen, Haifeng Liu, Xiaofei He, Binbin
  Lin, Wanli Ouyang, and Tong He.
\newblock Depth any video with scalable synthetic data.
\newblock {\em arXiv preprint arXiv:2410.10815}, 2024.

\bibitem{chen2025video}
Sili Chen, Hengkai Guo, Shengnan Zhu, Feihu Zhang, Zilong Huang, Jiashi Feng,
  and Bingyi Kang.
\newblock Video depth anything: Consistent depth estimation for super-long
  videos.
\newblock In {\em Proceedings of the Computer Vision and Pattern Recognition
  Conference}, pages 22831--22840, 2025.

\bibitem{he2024ma}
Bo~He, Hengduo Li, Young~Kyun Jang, Menglin Jia, Xuefei Cao, Ashish Shah,
  Abhinav Shrivastava, and Ser-Nam Lim.
\newblock Ma-lmm: Memory-augmented large multimodal model for long-term video
  understanding.
\newblock In {\em Proceedings of the IEEE/CVF conference on computer vision and
  pattern recognition}, pages 13504--13514, 2024.

\bibitem{Gammerman98learningby}
A.~Gammerman, V.~Vovk, and V.~Vapnik.
\newblock Learning by transduction.
\newblock In {\em In Uncertainty in Artificial Intelligence}, pages 148--155.
  Morgan Kaufmann, 1998.

\bibitem{vapnik_book}
Vladimir Vapnik and S.~Kotz.
\newblock {\em Estimation of Dependences Based on Empirical Data: Empirical
  Inference Science (Information Science and Statistics)}.
\newblock Springer-Verlag, Berlin, Heidelberg, 2006.

\bibitem{bottou1992local}
L{\'e}on Bottou and Vladimir Vapnik.
\newblock Local learning algorithms.
\newblock {\em Neural computation}, 4(6):888--900, 1992.

\bibitem{zhang2006svm}
Hao Zhang, Alexander~C Berg, Michael Maire, and Jitendra Malik.
\newblock Svm-knn: Discriminative nearest neighbor classification for visual
  category recognition.
\newblock In {\em 2006 IEEE Computer Society Conference on Computer Vision and
  Pattern Recognition (CVPR'06)}, volume~2, pages 2126--2136. IEEE, 2006.

\bibitem{hardt2023test}
Moritz Hardt and Yu~Sun.
\newblock Test-time training on nearest neighbors for large language models.
\newblock {\em arXiv preprint arXiv:2305.18466}, 2023.

\bibitem{qi2024safety}
Xiangyu Qi, Ashwinee Panda, Kaifeng Lyu, Xiao Ma, Subhrajit Roy, Ahmad Beirami,
  Prateek Mittal, and Peter Henderson.
\newblock Safety alignment should be made more than just a few tokens deep.
\newblock {\em arXiv preprint arXiv:2406.05946}, 2024.

\bibitem{jain2011online}
Vidit Jain and Erik Learned-Miller.
\newblock Online domain adaptation of a pre-trained cascade of classifiers.
\newblock In {\em CVPR 2011}, pages 577--584. IEEE, 2011.

\bibitem{shocher2018zero}
Assaf Shocher, Nadav Cohen, and Michal Irani.
\newblock “zero-shot” super-resolution using deep internal learning.
\newblock In {\em Proceedings of the IEEE Conference on Computer Vision and
  Pattern Recognition}, pages 3118--3126, 2018.

\bibitem{nitzan2022mystyle}
Yotam Nitzan, Kfir Aberman, Qiurui He, Orly Liba, Michal Yarom, Yossi
  Gandelsman, Inbar Mosseri, Yael Pritch, and Daniel Cohen-Or.
\newblock Mystyle: A personalized generative prior.
\newblock {\em arXiv preprint arXiv:2203.17272}, 2022.

\bibitem{xie2023sepico}
Binhui Xie, Shuang Li, Mingjia Li, Chi~Harold Liu, Gao Huang, and Guoren Wang.
\newblock Sepico: Semantic-guided pixel contrast for domain adaptive semantic
  segmentation.
\newblock {\em IEEE Transactions on Pattern Analysis and Machine Intelligence},
  2023.

\bibitem{tonioni2019learning}
Alessio Tonioni, Oscar Rahnama, Thomas Joy, Luigi~Di Stefano, Thalaiyasingam
  Ajanthan, and Philip~HS Torr.
\newblock Learning to adapt for stereo.
\newblock In {\em Proceedings of the IEEE/CVF Conference on Computer Vision and
  Pattern Recognition}, pages 9661--9670, 2019.

\bibitem{tonioni2019real}
Alessio Tonioni, Fabio Tosi, Matteo Poggi, Stefano Mattoccia, and Luigi~Di
  Stefano.
\newblock Real-time self-adaptive deep stereo.
\newblock In {\em Proceedings of the IEEE/CVF Conference on Computer Vision and
  Pattern Recognition}, pages 195--204, 2019.

\bibitem{zhang2020online}
Zhenyu Zhang, Stephane Lathuiliere, Elisa Ricci, Nicu Sebe, Yan Yan, and Jian
  Yang.
\newblock Online depth learning against forgetting in monocular videos.
\newblock In {\em Proceedings of the IEEE/CVF Conference on Computer Vision and
  Pattern Recognition}, pages 4494--4503, 2020.

\bibitem{zhong2018open}
Yiran Zhong, Hongdong Li, and Yuchao Dai.
\newblock Open-world stereo video matching with deep rnn.
\newblock In {\em Proceedings of the European Conference on Computer Vision
  (ECCV)}, pages 101--116, 2018.

\bibitem{luo2020consistent}
Xuan Luo, Jia-Bin Huang, Richard Szeliski, Kevin Matzen, and Johannes Kopf.
\newblock Consistent video depth estimation.
\newblock {\em ACM Transactions on Graphics (ToG)}, 39(4):71--1, 2020.

\bibitem{hansen2020self}
Nicklas Hansen, Rishabh Jangir, Yu~Sun, Guillem Aleny{\`a}, Pieter Abbeel,
  Alexei~A Efros, Lerrel Pinto, and Xiaolong Wang.
\newblock Self-supervised policy adaptation during deployment.
\newblock {\em arXiv preprint arXiv:2007.04309}, 2020.

\bibitem{sun2021online}
Yu~Sun, Wyatt~L Ubellacker, Wen-Loong Ma, Xiang Zhang, Changhao Wang, Noel~V
  Csomay-Shanklin, Masayoshi Tomizuka, Koushil Sreenath, and Aaron~D Ames.
\newblock Online learning of unknown dynamics for model-based controllers in
  legged locomotion.
\newblock {\em IEEE Robotics and Automation Letters}, 6(4):8442--8449, 2021.

\bibitem{liu2021ttt++}
Yuejiang Liu, Parth Kothari, Bastien van Delft, Baptiste Bellot-Gurlet, Taylor
  Mordan, and Alexandre Alahi.
\newblock Ttt++: When does self-supervised test-time training fail or thrive?
\newblock {\em Advances in Neural Information Processing Systems}, 34, 2021.

\bibitem{yuan2023robust}
Longhui Yuan, Binhui Xie, and Shuang Li.
\newblock Robust test-time adaptation in dynamic scenarios.
\newblock In {\em Proceedings of the IEEE/CVF Conference on Computer Vision and
  Pattern Recognition}, pages 15922--15932, 2023.

\bibitem{volpi2022road}
Riccardo Volpi, Pau De~Jorge, Diane Larlus, and Gabriela Csurka.
\newblock On the road to online adaptation for semantic image segmentation.
\newblock In {\em Proceedings of the IEEE/CVF Conference on Computer Vision and
  Pattern Recognition}, pages 19184--19195, 2022.

\bibitem{wiegand2003overview}
Thomas Wiegand, Gary~J Sullivan, Gisle Bjontegaard, and Ajay Luthra.
\newblock Overview of the h. 264/avc video coding standard.
\newblock {\em IEEE Transactions on circuits and systems for video technology},
  13(7):560--576, 2003.

\bibitem{sutton1995generalization}
Richard~S Sutton.
\newblock Generalization in reinforcement learning: Successful examples using
  sparse coarse coding.
\newblock {\em Advances in neural information processing systems}, 8, 1995.

\bibitem{hinton2022forward}
Geoffrey Hinton.
\newblock The forward-forward algorithm: Some preliminary investigations.
\newblock {\em arXiv preprint arXiv:2212.13345}, 2(3):5, 2022.

\bibitem{lowe2019putting}
Sindy L{\"o}we, Peter O'Connor, and Bastiaan Veeling.
\newblock Putting an end to end-to-end: Gradient-isolated learning of
  representations.
\newblock {\em Advances in neural information processing systems}, 32, 2019.

\bibitem{lee2015difference}
Dong-Hyun Lee, Saizheng Zhang, Asja Fischer, and Yoshua Bengio.
\newblock Difference target propagation.
\newblock In {\em Joint european conference on machine learning and knowledge
  discovery in databases}, pages 498--515. Springer, 2015.

\bibitem{li2025noprop}
Qinyu Li, Yee~Whye Teh, and Razvan Pascanu.
\newblock Noprop: Training neural networks without full back-propagation or
  full forward-propagation.
\newblock {\em arXiv preprint arXiv:2503.24322}, 2025.

\bibitem{hinton2021glomgodfather}
Cade Metz.
\newblock Geoffrey hinton—the ‘godfather’ of ai and neural networks.
\newblock {\em MIT Technology Review}, 2021.

\bibitem{hinton1995wake}
Geoffrey~E Hinton, Peter Dayan, Brendan~J Frey, and Radford~M Neal.
\newblock The" wake-sleep" algorithm for unsupervised neural networks.
\newblock {\em Science}, 268(5214):1158--1161, 1995.

\bibitem{hinton1984boltzmann}
Geoffrey~E Hinton, Terrence~J Sejnowski, and David~H Ackley.
\newblock {\em Boltzmann machines: Constraint satisfaction networks that
  learn}.
\newblock Carnegie-Mellon University, Department of Computer Science
  Pittsburgh, PA, 1984.

\bibitem{song2020score}
Yang Song, Jascha Sohl-Dickstein, Diederik~P Kingma, Abhishek Kumar, Stefano
  Ermon, and Ben Poole.
\newblock Score-based generative modeling through stochastic differential
  equations.
\newblock {\em arXiv preprint arXiv:2011.13456}, 2020.

\bibitem{sun2024learning}
Yu~Sun, Xinhao Li, Karan Dalal, Jiarui Xu, Arjun Vikram, Genghan Zhang, Yann
  Dubois, Xinlei Chen, Xiaolong Wang, Sanmi Koyejo, et~al.
\newblock Learning to (learn at test time): Rnns with expressive hidden states.
\newblock {\em arXiv preprint arXiv:2407.04620}, 2024.

\bibitem{dalal2025one}
Karan Dalal, Daniel Koceja, Gashon Hussein, Jiarui Xu, Yue Zhao, Youjin Song,
  Shihao Han, Ka~Chun Cheung, Jan Kautz, Carlos Guestrin, et~al.
\newblock One-minute video generation with test-time training.
\newblock {\em arXiv preprint arXiv:2504.05298}, 2025.

\bibitem{cheng2022mask2former}
Bowen Cheng, Ishan Misra, Alexander~G Schwing, Alexander Kirillov, and Rohit
  Girdhar.
\newblock {Masked-attention Mask Transformer for Universal Image Segmentation}.
\newblock In {\em CVPR}, 2022.

\bibitem{soomro2012ucf101}
Khurram Soomro, Amir~Roshan Zamir, and Mubarak Shah.
\newblock Ucf101: A dataset of 101 human actions classes from videos in the
  wild.
\newblock {\em arXiv preprint arXiv:1212.0402}, 2012.

\bibitem{cordts2016cityscapes}
Marius Cordts, Mohamed Omran, Sebastian Ramos, Timo Rehfeld, Markus Enzweiler,
  Rodrigo Benenson, Uwe Franke, Stefan Roth, and Bernt Schiele.
\newblock The cityscapes dataset for semantic urban scene understanding.
\newblock In {\em Proceedings of the IEEE conference on computer vision and
  pattern recognition}, pages 3213--3223, 2016.

\bibitem{he2022masked}
Kaiming He, Xinlei Chen, Saining Xie, Yanghao Li, Piotr Doll{\'a}r, and Ross
  Girshick.
\newblock Masked autoencoders are scalable vision learners.
\newblock In {\em Proceedings of the IEEE/CVF conference on computer vision and
  pattern recognition}, pages 16000--16009, 2022.

\bibitem{gandelsman2022test}
Yossi Gandelsman, Yu~Sun, Xinlei Chen, and Alexei Efros.
\newblock Test-time training with masked autoencoders.
\newblock {\em Advances in Neural Information Processing Systems},
  35:29374--29385, 2022.

\end{thebibliography}

\newpage

\vspace{10em}
\noindent \textbf{\Large Table of Contents}
\vspace{1em}
\hrule
\vspace{4em}

\noindent

\textbf{A \quad Broader Impact} \dotfill 
\textbf{\pageref{sec:broader_impact}}\\

\textbf{B \quad Reproducibility Statement} \dotfill 
\textbf{\pageref{sec:reproducibility}}\\

\textbf{C \quad Additional Ablations} \dotfill 
\textbf{\pageref{sec:supp_ablations}}\\

\textbf{D \quad Details of Datasets} \dotfill 
\textbf{\pageref{sec:supp_datasets}}\\

\textbf{E \quad A brief overview of architectures} \dotfill 
\textbf{\pageref{sec:supp_architecture}}\\

\textbf{F \quad Qualitative Demonstrations of EpicTours Dataset} \dotfill 
\textbf{\pageref{fig:epictours_dataset_1} - \pageref{fig:epictours_dataset_3}}\\
\vspace{4em}
\hrule

\newpage

\section{Broader Impact}
\label{sec:broader_impact}

This research introduces the Frame Forgetting Network (FFN) for efficient test-time training on long videos, along with the EpicTours dataset for benchmarking adaptation on multi-hour video streams. FFN demonstrates that test-time training can be made computationally tractable for hours-long videos by operating on \textit{only} three frames per sliding window transition. This may benefit the real-world deployment of video understanding systems in resource-constrained environments, such as edge devices, drones, and mobile platforms.\\

\noindent Similarly, by removing the dependency on a remote teacher model, FFN supports \textit{fully} on-device adaptation. This is particularly relevant for offline scenarios such as disaster-relief efforts, autonomous navigation, and remote surveillance, where server connectivity may be limited or unavailable. Our adaptive windowing algorithm (AWA) reduces \textit{unnecessary} computation by dynamically deciding \textit{when} adaptation is needed, rather than performing TTT on \textit{every} frame. This may help contribute to more energy-efficient video processing pipelines.\\

 \noindent  We do not anticipate direct negative social impacts from this work, beyond what existing on-device models already face. However, we acknowledge that video understanding technologies, including those built on TTT, could potentially be applied in surveillance contexts. We encourage the responsible use of such technologies, in accordance with applicable regulations and ethical guidelines.

\section{Reproducibility Statement}
\label{sec:reproducibility}

To ensure the reproducibility of our experiments, we provide a comprehensive overview of the model architectures, hyperparameters, evaluation procedures, datasets, and baselines employed in this supplementary material. We provide complete dataset details in \cref{sec:supp_datasets} and architecture details in \cref{sec:supp_architecture}. Our code and data will be made publicly available for future research purposes. The videos will be made available under the Creative Commons (CC) license. \\

\noindent For segmentation experiments (Tab.~1, Tab.~5), we initialize the backbone $f$ with Mask2Former~\cite{cheng2022mask2former} pre-trained weights, train the SSL head $g$ from scratch \textit{jointly} during the training-time phase, and use Mask2Former's pre-trained weights for the downstream head $h$. For depth estimation (Tab.~3), we leverage the Video Depth Anything architecture~\cite{chen2025video}. For video classification (Tab.~4), we adopt the DINOv3~\cite{simeoni2025dinov3} backbone. \\

\noindent Across all experiments, we set the memory buffer size $W = 50$, perform a \textit{single} gradient step per incoming frame, and use the Adam optimizer. The SSL head $g$ is trained with the \textit{next-frame anticipation} objective instead of the current-frame reconstruction (as discussed in the main paper and validated in Fig.~5(iii)). We perform all experiments with 3 random seeds and report their mean accuracy. All experiments are conducted on a single NVIDIA Ampere (80GB) GPU. \\ 

\noindent During the \textit{training-time} phase, $g$ may be trained on both image datasets and a single long video; the anticipative head itself can be learned with as few as one long video. At \textit{test time}, the downstream head $h$ is \textit{kept frozen} throughout, and \textit{only} the backbone $f$ and SSL head $g$ receive gradient updates. The adaptive windowing algorithm (AWA) uses the surprise metric $S_t$ and the dynamic threshold $\tau_t = \mu_t + \sigma_t$ (Eq.~7 of the main paper) to \textit{dynamically} determine whether TTT is performed on a given frame. \\

\section{Additional Ablations}
\label{sec:supp_ablations}
Here, we provide some additional ablations for our FFN. 

\begin{table}[h]
\vspace{-2em}
\centering
\caption{\textbf{Ablation on  varying number of MLP layers in temporal module}. We report results for \textit{instance segmentation} on COCO-Videos dataset. Higher is better.}
\label{tab:mlp_layers}
\setlength{\tabcolsep}{10pt} 
\begin{tabular}{@{}l|cccc@{}}
\toprule
Layers & 1 & 3 & 5 & 10 \\ \midrule
 & 41.2 & \textbf{45.1} & 43.8 & 39.5 \\ \bottomrule
\end{tabular}
\label{tab:varying_mlp_layers}
\vspace{-1.5em}
\end{table}

\noindent \textbf{Varying the number of MLP layers in the temporal module:} Tab \ref{tab:varying_mlp_layers} reveals that optimal number of layers for temporal module is $3$.

\vspace{1em}

\noindent \textbf{Positional encodings \textit{do not} repeat even if sampled for very long videos:} Another criticism may be that FFN relies on temporally binding (a.k.a conditioning) the backbone with different timesteps. Since the positional encodings are circular, eventually they might end up repeating, which would mean that the network could no longer distinguish between timsesteps of different indices. We refer to  this problem as the `temporal symmetry problem'.

\vspace{1em}
\lstdefinestyle{pythonstyle}{
    language=Python,
    backgroundcolor=\color{gray!5},
    basicstyle=\ttfamily\small,
    keywordstyle=\color{blue!70!black},
    commentstyle=\color{gray},
    stringstyle=\color{green!50!black},
    numbers=left,
    numberstyle=\tiny\color{gray},
    stepnumber=1,
    numbersep=8pt,
    showstringspaces=false,
    breaklines=true,
    frame=lines,
    captionpos=b
}

\begin{lstlisting}[style=pythonstyle, caption={Our Implementation of checking collisions in positional encodings.}, label={lst:positional_hash}]
import numpy as np
import hashlib
import collections

def get_1d_positional_encoding_np(t, d, lambda_=10000):
    """Computes sinusoidal positional encoding for a time step."""
    half_d = int(np.ceil(d / 2))
    indices = np.arange(half_d)
    inv_freq = 1.0 / (lambda_ ** (indices / (d / 2)))
    sin_inp = t * inv_freq
    emb = np.concatenate([np.sin(sin_inp), np.cos(sin_inp)])
    return emb[:d].astype(np.float32)

def get_full_float_hash(vec):
    """Computes a unique 64-bit signature from raw float memory."""
    raw_bytes = np.ascontiguousarray(vec).tobytes()
    h = hashlib.blake2b(raw_bytes, digest_size=8)
    return int(h.hexdigest(), 16)

# Configuration for temporal windows
total_time = [60, 3600, 86400, 2592000]
wavelengths = [500, 5000, 10000, 20000, 50000]
all_hashes = []

for w in wavelengths:
    for time in total_time:
        signature_hash = collections.defaultdict(int)
        for t in range(time):
            pos = get_1d_positional_encoding_np(t=t, d=2, lambda_=w)
            h = get_full_float_hash(np.sort(pos))
            signature_hash[h] += 1
        all_hashes.append({'wavelength': w, 'time': time, 'hash': signature_hash})
\end{lstlisting}
\vspace{-0.5em}

\noindent In the code above, we implement 1D absolute positional encodings. For a particular timestep, we return a $d$ dimensional vector, and subsequently hash it to an integer. Simultaneously, we track potential collisions in the hash in case two different timesteps led to the same encodings. \\

\noindent \textbf{Key observation:} We observe that this formulation \textit{does not} suffer from any collisons, thereby indicating that the neural network is always given \textit{unique} inputs for different time indices. We also test extreme case by reducing the dimensionality $d$ of positional encoding to $2$ (one sin, another cosine) and found that it still broke symmetry. This means that the cyclic nature of sines, and cosines, can still work. This might also help validate the idea that a particular signal can be written as a superposition of sines and cosines (fourier's expansion)

\section{Datasets}
\label{sec:supp_datasets}

Next, we describe the details of every dataset used in our experiments.\\

\noindent \textbf{COCO-Videos} COCO-Videos~\cite{wang2025test} is a video dataset derived from the COCO benchmark, designed to evaluate test-time training in video streams. It contains approximately 30k frames with videos averaging around 309 seconds in length. The dataset provides \textit{dense} annotations suitable for evaluating both instance segmentation and panoptic segmentation. \\

\noindent \textbf{KITTI-STEP:} KITTI-STEP~\cite{weber2021step} extends the KITTI benchmark with dense, \textit{temporally consistent} panoptic segmentation annotations. It contains approximately 8k frames with video sequences averaging 40 seconds in length. The dataset includes both a validation and a test split. We report on semantic segmentation performance in terms of mIoU in both splits. KITTI-STEP is particularly relevant to our evaluation because its driving sequences exhibit gradual scene transitions; this allows us to study how FFN's principle of locality and adaptive windowing handle \textit{slowly} evolving visual content, as opposed to the more abrupt scene changes found in other benchmarks.\\

\noindent \textbf{UCF101:} UCF101~\cite{soomro2012ucf101} is a widely-used action recognition benchmark containing 13,320 video clips spanning 101 action categories. The videos are collected from YouTube and cover a diverse range of human actions. We report top-1 classification accuracy on this dataset. While our primary focus is on dense per-pixel tasks like segmentation, UCF101 allows us to evaluate whether FFN's adaptation mechanism can also benefit \textit{coarse} video understanding. \\

\noindent \textbf{Something-Something v2}:(SSv2)~\cite{goyal2017something} is a video understanding dataset that requires \textit{temporal reasoning} about object interactions. It contains more than 220k short video clips spanning 174 fine-grained action categories. Notably, SSv2 requires models to reason about the \textit{direction of time} (\eg, ``moving something up'' vs.\ ``moving something down''), making it a challenging benchmark for temporal understanding. We report top-1 classification accuracy. SSv2 is especially interesting for FFN because it tests whether our anticipative SSL objective, which predicts the \textit{next} frame rather than reconstructing the current one, can capture the temporal directionality that this benchmark demands.\\

\noindent \textbf{ScanNet}: ScanNet~\cite{dai2017scannet} is a richly-annotated dataset of 3D indoor scene reconstructions. It contains RGB-D video sequences captured in a variety of indoor environments. We use ScanNet to evaluate the video depth estimation, reporting AbsRel and $\delta_1$ metrics. Following prior work~\cite{chen2025video}, we also report the Temporal Alignment Error (TAE) on a 170-frame variant of ScanNet. The TAE metric is particularly informative for FFN, as it measures \textit{temporal consistency} of depth predictions across frames; this directly tests whether FFN's memory restoration mechanism and adaptive windowing can maintain \textit{coherent} depth estimates over time.\\

\noindent \textbf{Bonn:} Bonn~\cite{palazzolo2019iros} is a dataset designed for 3D reconstruction in dynamic environments using RGB-D cameras. It provides ground-truth depth annotations for indoor scenes. We use Bonn to evaluate video depth estimation and report AbsRel and $\delta_1$ metrics. Unlike ScanNet and NYUv2, Bonn specifically targets \textit{dynamic} indoor environments where objects and people may move throughout the scene. This makes it a useful test of whether FFN's adaptation can handle \textit{non-static} visual content during depth estimation.\\

\noindent \textbf{NYUv2}: NYUv2~\cite{Silberman:ECCV12} is a widely-used indoor depth estimation benchmark containing RGB-D images captured with a Microsoft Kinect sensor. It provides dense depth annotations for a variety of indoor scenes. We evaluate video depth estimation on NYUv2 and report AbsRel and $\delta_1$ metrics. NYUv2 covers a broad range of room layouts and furniture configurations, making it a robust generalization test. Together with KITTI (outdoor), ScanNet (indoor 3D), and Bonn (dynamic indoor), NYUv2 rounds out a diverse evaluation suite that allows us to assess FFN's depth estimation across \textit{varied} environments and sensor characteristics.\\

\noindent \textbf{Sintel:} Sintel~\cite{Butler_Wulff_Stanley_Black_2012} is a synthetic benchmark derived from the open-source animated film \textit{Sintel}. It provides ground-truth depth and optical flow annotations. Following prior work, we evaluate a 50-frame variant and report AbsRel and $\delta_1$ metrics for video depth estimation. The synthetic nature of Sintel introduces a significant \textit{domain gap} relative to the real-world datasets above; strong performance here may indicate that FFN's self-supervised adaptation can bridge the gap between a model's training domain and previously unseen visual styles.\\

\noindent \textbf{CityScapes:} CityScapes~\cite{cordts2016cityscapes} is a large-scale dataset to understand semantic urban scenes, containing video sequences captured from a vehicle driving through various cities. It provides fine-grained pixel-level annotations for 19 semantic classes. As shown in Tab.~2 of the main paper, CityScapes videos average only 1.8 seconds with approximately 3k frames, making it representative of the \textit{short-video} regime that our EpicTours dataset aims to extend. We include CityScapes in Tab.~2 primarily to contextualize the scale gap: while existing benchmarks operate on seconds-long clips, FFN is designed for videos that are \textit{orders of magnitude} longer.\\

\noindent \textbf{DAVIS:} DAVIS~\cite{Perazzi2016davis} is a benchmark dataset for video object segmentation, containing densely annotated video sequences averaging 3.5 seconds with approximately 3.4k frames. It is widely used to evaluate methods on short video clips and serves as a reference point in Tab.~2 to compare video lengths across datasets. Although DAVIS provides high-quality per-frame annotations, its short clip lengths mean that temporal adaptation methods like FFN have limited opportunity to demonstrate their long-range benefits on this benchmark alone.

\section{Architecture}
\label{sec:supp_architecture}

\subsection{Mask2Former}
\label{sec:arch_mask2former}

Our segmentation experiments employ Mask2Former~\cite{cheng2022mask2former} as the backbone architecture. Mask2Former is a \textit{universal} architecture for image segmentation that unifies semantic, instance, and panoptic segmentation under a single framework. It consists of three main components: (i)~a pixel-level feature extraction backbone, (ii)~a Transformer-based decoder with masked attention, and (iii)~a set of learnable object queries that produce per-segment predictions. The key innovation of Mask2Former is its \textit{masked cross-attention} mechanism, which restricts attention to localized regions around predicted segments rather than the full image, improving both efficiency and segmentation quality. \\

\noindent In our FFN framework, we use Mask2Former's pre-trained backbone as $f$ and its pre-trained task-specific head as $h$. The SSL head $g$ is trained \textit{from scratch}. During the training-time phase, $(f, g, h)$ are trained \textit{jointly}. During test-time training, only $f$ and $g$ are updated, while $h$ is \textit{kept frozen}.

\subsection{TTT-MAE: Test-Time Training with Masked Autoencoders}
\label{sec:arch_ttt_mae}

Our self-supervised learning (SSL) task for test-time training is based on the Masked Autoencoder (MAE) framework~\cite{he2022masked, gandelsman2022test}. MAE operates by randomly masking a large proportion of input image patches and training the model to reconstruct the missing patches. This reconstruction task provides a strong self-supervised signal that captures both local and global image structure. In the TTT-MAE setup~\cite{gandelsman2022test}, the MAE reconstruction objective serves as the self-supervised task during \textit{both} the training-time and test-time phases. Specifically, given an incoming frame $x_t$, the SSL head $g$ takes the backbone features $f(x_t)$ and produces a reconstruction $x'_t = g \circ f(x_t)$. The reconstruction loss $\ell_s(x'_t, x_t)$ is then used to update the backbone $f$ and SSL head $g$ via backpropagation. \\

\noindent The key advantage of using MAE as the SSL task is that it does not require any external labels, making it well-suited for test-time training where ground-truth annotations are unavailable. In our FFN, we replace the standard current-frame reconstruction objective with a \textit{next-frame anticipation} objective, where the SSL head is trained to predict the upcoming frame rather than reconstruct the current one.

\begin{figure*}[t]
  \centering
  \includegraphics[width=\textwidth]{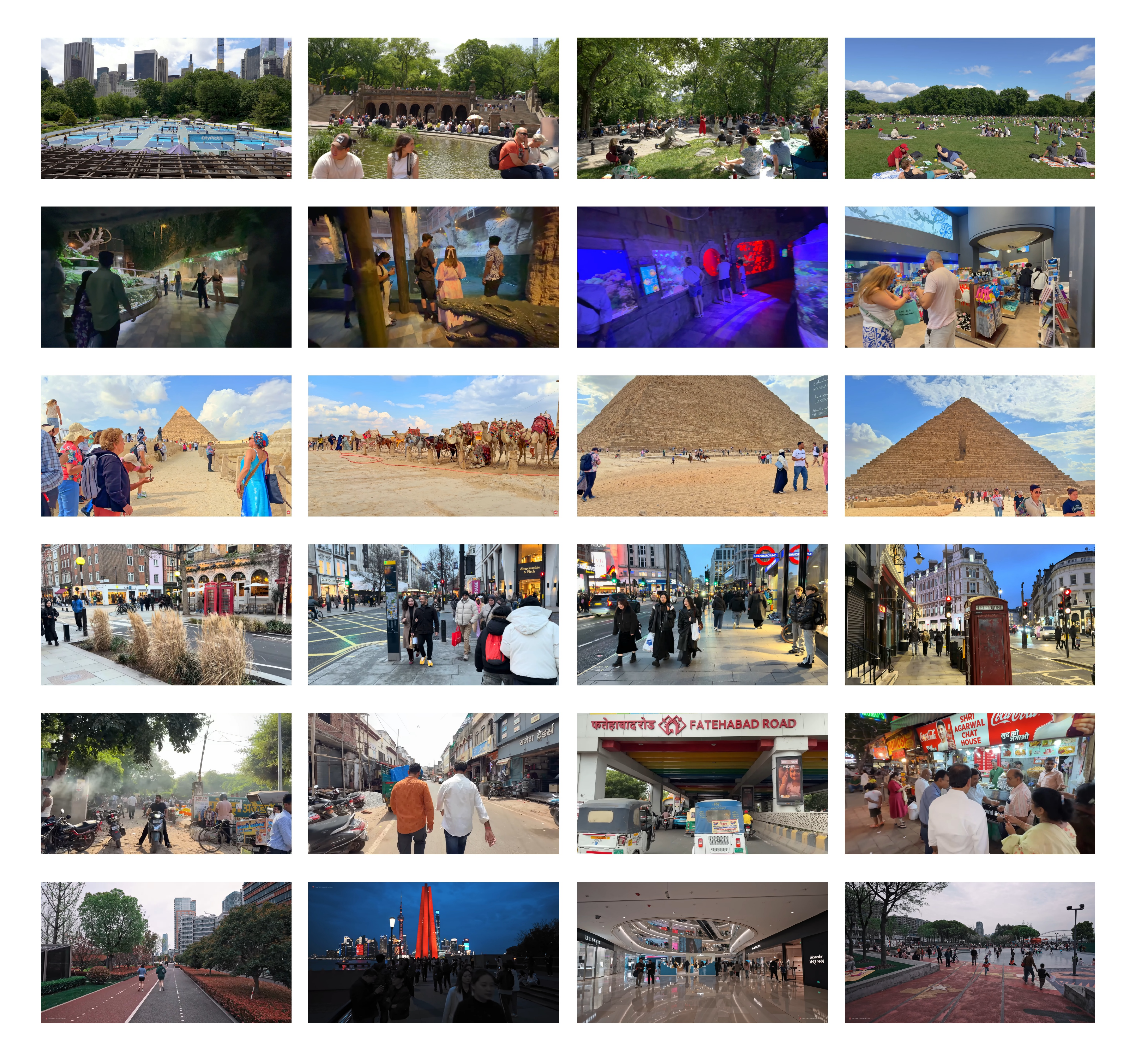}
  \caption{\textbf{Diversity of our EpicTours Dataset}: Each row contains different videos, different columns contain frames in each video. Best viewed in color.}
  \label{fig:epictours_dataset_1}
\end{figure*}
\clearpage

\begin{figure*}[t]
  \centering
  \includegraphics[width=\textwidth]{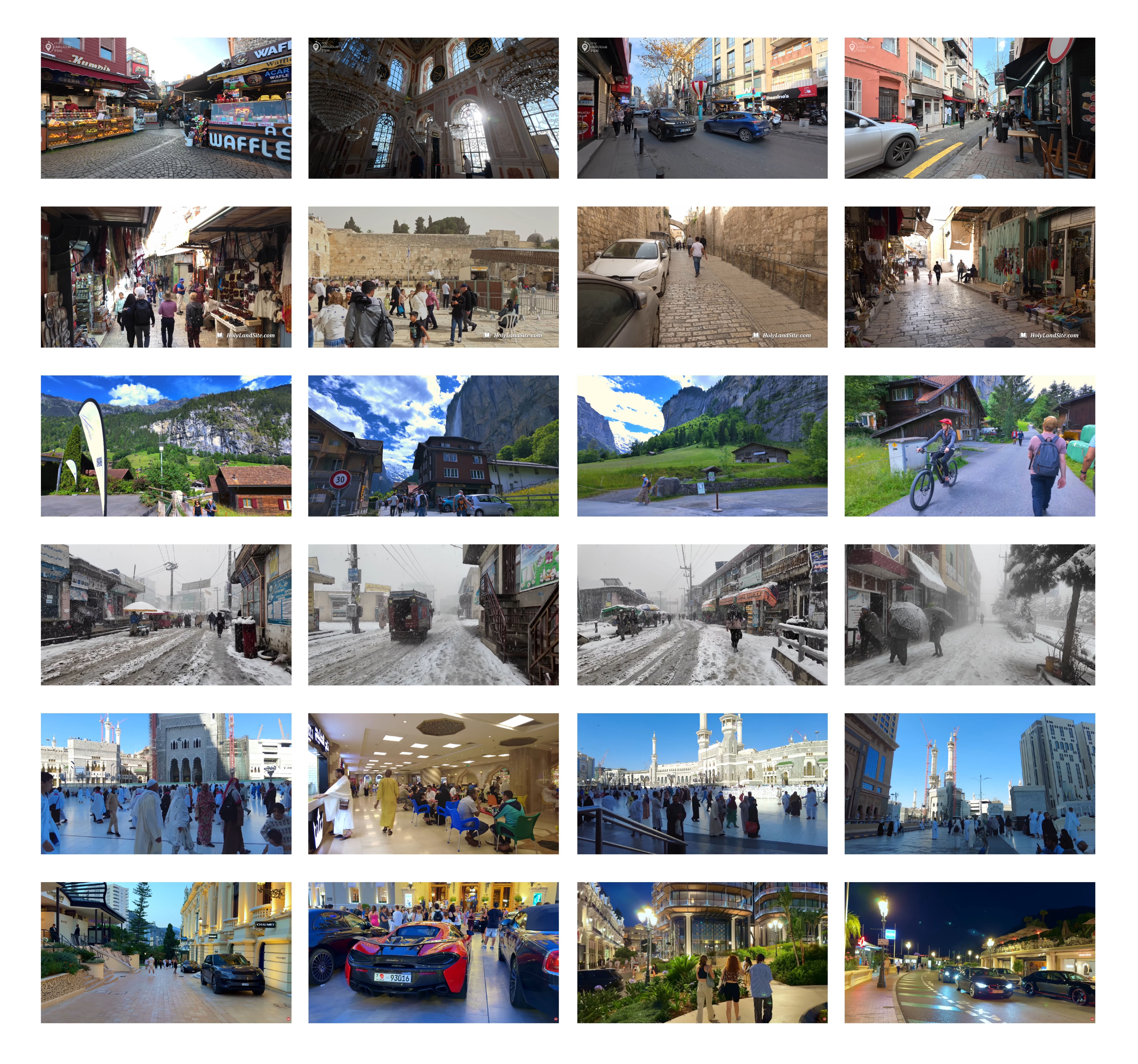}
  \caption{\textbf{Diversity of our EpicTours Dataset}: Each row contains different videos, different columns contain frames in each video. Best viewed in color.}
  \label{fig:epictours_dataset_2}
\end{figure*}
\clearpage

\begin{figure*}[t]
  \centering
  \includegraphics[width=\textwidth]{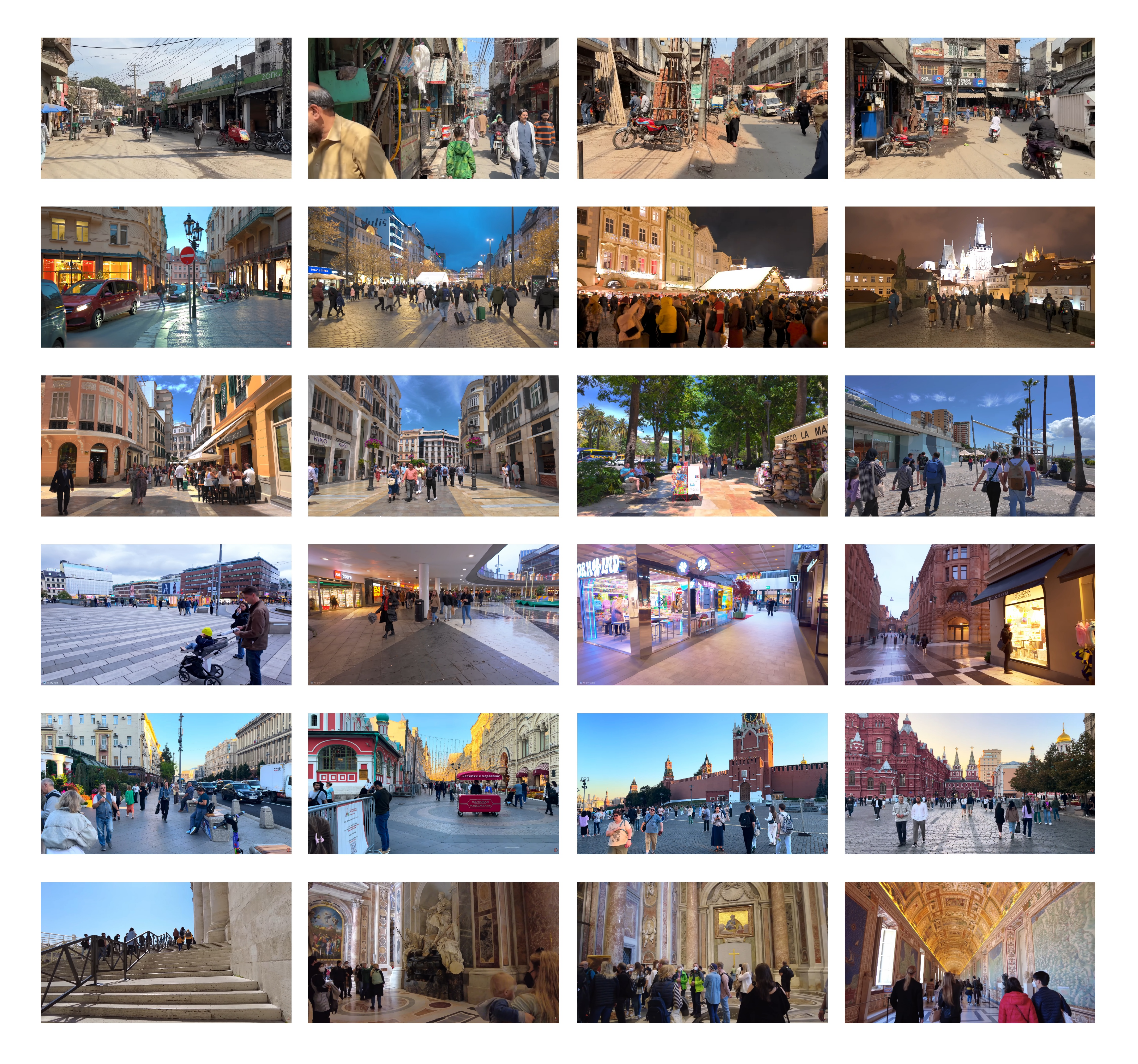}
  \caption{\textbf{Diversity of our EpicTours Dataset}: Each row contains different videos, different columns contain frames in each video. Best viewed in color.}
  \label{fig:epictours_dataset_3}
\end{figure*}
\clearpage

\end{document}